\documentclass{firered}
\usepackage[utf8]{inputenc} 
\usepackage[T1]{fontenc}    
\usepackage{hyperref}       
\usepackage{url}            
\usepackage{booktabs}       
\usepackage{amsfonts}       
\usepackage{nicefrac}       
\usepackage{microtype}      
\usepackage{float}
\usepackage{mwe}
\usepackage{graphicx}
\usepackage{amssymb}
\usepackage{adjustbox}
\usepackage{colortbl}
\usepackage{array}
\usepackage{multirow}
\usepackage{makecell}
\usepackage{bbm}
\usepackage{collcell,xfp}
\usepackage{pgf}
\usepackage{tikz}
\usepackage[most]{tcolorbox}
\usepackage{csquotes}
\usepackage[noorphans,vskip=1em,leftmargin=1em]{quoting}
\usepackage{pifont} 
\usepackage{enumitem} 
\usepackage{tcolorbox} 
\usepackage{forest}
\usepackage{wrapfig}
\usepackage{caption}
\usepackage{subcaption}
\usepackage{longtable}
\usepackage{algpseudocode}
\usepackage{amsmath}
\usepackage[capitalize]{cleveref}
\usepackage[ruled,vlined,linesnumbered]{algorithm2e} 
\usepackage[percent]{overpic}
\usepackage{xspace}
\usepackage[normalem]{ulem}  
\usepackage{tocloft}  
\usepackage{fancyhdr} 
\usepackage{natbib}  

\fancypagestyle{firststyle}{
    \fancyhead[L]{}
    \fancyhead[C]{}
    \fancyhead[R]{}
    
    \renewcommand{\footrulewidth}{0.8pt}
    \renewcommand{\headrule}{}
    \renewcommand{\footrule}{\color{sectionred}\hrule width\textwidth height\footrulewidth \vspace{-100pt}}
    \fancyfoot[C]{}
}

\renewcommand{\headrule}{}
\renewcommand{\footrulewidth}{0pt}

\usepackage{listings}
\usepackage{wrapfig}
\usepackage{geometry}
\usepackage{xcolor}
\usepackage[most]{tcolorbox}

\usepackage{CJKutf8}   

\usepackage{twemojis}
\usepackage{tikz-dependency}
\usepackage{graphicx}
\usepackage{tikz}
\usepackage{tikz-qtree}
\usepackage{caption}
\usetikzlibrary{arrows.meta}
\definecolor{clearpurple}{RGB}{138, 140, 191}
\definecolor{clearyellow}{HTML}{f2d3bf}
\definecolor{skyblue}{HTML}{a8d8e2}
\definecolor{darkblue}{HTML}{19183B}
\definecolor{tp1}{RGB}{253, 207, 158}

\usepackage{fontawesome}
\newcommand{\huggingface}{\raisebox{-1.5pt}{\includegraphics[height=1.05em]{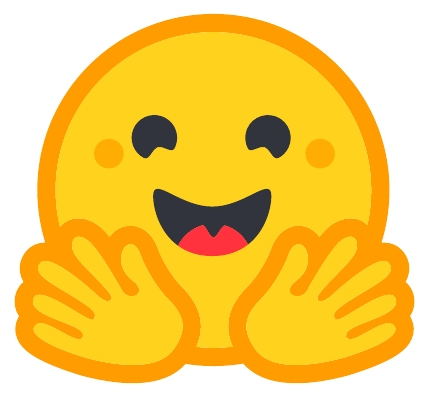}}\xspace}

\newcommand{\github}{\raisebox{-1.5pt}{\includegraphics[height=1.05em]{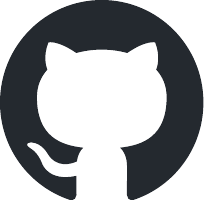}}\xspace}

\definecolor{scholarblue}{rgb}{0.21,0.49,0.74}
\definecolor{bluelink}{RGB}{0,113,188}
\definecolor{greenlink}{RGB}{0,188,113}
\hypersetup{
    colorlinks=true,%
    citecolor=scholarblue,%
    filecolor=red,%
    linkcolor=red!93!black,%
    urlcolor=bluelink
}

\newtcolorbox{promptbox}[1][]{
    enhanced,
    colback=gray!10,      
    colframe=black,       
    coltitle=white,       
    fonttitle=\bfseries\sffamily, 
    fontupper=\ttfamily\small,    
    boxrule=0.8pt,        
    arc=4mm,              
    title={#1},           
    attach boxed title to top left={xshift=0.5cm, yshift*=-\tcboxedtitleheight/2},
    boxed title style={
        colback=black,    
        arc=3mm,          
        boxrule=0pt,      
        left=3mm, right=3mm, top=1.5mm, bottom=1.5mm
    },
    top=1.5em,            
    parbox=false          
}

\usepackage{titlesec}
\titlespacing*{\paragraph}
    {0pt}     
    {0.25em}  
    {1em}  

\definecolor{sectionred}{RGB}{220,38,38}  
\titleformat{\section}{\color{sectionred}\large\bfseries}{\color{sectionred}\thesection.}{0.5em}{#1}[]
\titleformat{\subsection}{\color{sectionred}\bfseries}{\color{sectionred}\thesubsection.}{0.5em}{#1}[]
\titleformat{\subsubsection}{\color{sectionred}\bfseries\itshape}{\color{sectionred}\thesubsubsection.}{0.5em}{#1}[]

\definecolor{cGreen}{RGB}{69, 139, 89}    
\definecolor{cYellow}{RGB}{241, 186, 88}  
\definecolor{cBlue}{RGB}{95, 126, 222}    
\definecolor{cRed}{RGB}{219, 86, 96}      
\definecolor{cPurple}{RGB}{166, 156, 226} 
\definecolor{gridColor}{RGB}{200, 200, 200}
\definecolor{bgColor}{RGB}{248, 252, 252}

\definecolor{navyblue}{HTML}{0071BC}
\iftrue   
\newcommand{\displaytodo}[1]{#1}
\else
\newcommand{\displaytodo}[1]{}
\fi

\newcommand{\methodname}{{FireRed-Image-Edit }}

\definecolor{blindcolor}{HTML}{AB2AC6}    
\definecolor{chancecolor}{HTML}{F59E0B}   
\definecolor{singlecolor}{HTML}{06B6D4}   
\definecolor{multiplecolor}{HTML}{2563EB} 
\definecolor{captioncolor}{HTML}{22C55E}  

 \newcommand{\culine}[2]{%
    \def\temp@uline{\bgroup\markoverwith
        {\textcolor{#1}{\rule[-0.5ex]{2pt}{1pt}}}\ULon}%
    \temp@uline{#2}%
}
 \newcommand{\cthickuline}[3][0.8pt]{%
    \def\temp@uline{\bgroup\markoverwith
        {\textcolor{#2}{\rule[-0.5ex]{2pt}{#1}}}\ULon}%
    \temp@uline{#3}%
}

\title{\center{FireRed-Image-Edit-1.0 Technical Report}} 

\renewcommand\Affilfont{\normalfont\fontsize{11}{15}\selectfont\centering}
\author{ 
    Super Intelligence Team, Xiaohongshu Inc.
}

\begin{document}

{\color{sectionred}\hrule width\textwidth height0.2pt}

\renewcommand{\abstractname}{\textcolor{sectionred}{Abstract}}
\begin{abstract}
We present FireRed-Image-Edit, a diffusion transformer for instruction-based image editing that achieves state-of-the-art performance through systematic optimization of data curation, training methodology, and evaluation design.
We construct a 1.6B-sample training corpus, comprising 900M text-to-image and 700M image editing pairs from diverse sources. After rigorous cleaning, stratification, auto-labeling, and two-stage filtering, we retain over 100M high-quality samples balanced between generation and editing, ensuring strong semantic coverage and instruction alignment.
Our multi-stage training pipeline progressively builds editing capability via pre-training, supervised fine-tuning, and reinforcement learning. To improve data efficiency, we introduce a Multi-Condition Aware Bucket Sampler for variable-resolution batching and Stochastic Instruction Alignment with dynamic prompt re-indexing. To stabilize optimization and enhance controllability, we propose Asymmetric Gradient Optimization for DPO, DiffusionNFT with layout-aware OCR rewards for text editing, and a differentiable Consistency Loss for identity preservation.
We further establish REDEdit-Bench, a comprehensive benchmark spanning 15 editing categories, including newly introduced beautification and low-level enhancement tasks. Extensive experiments on REDEdit-Bench and public benchmarks (ImgEdit and GEdit) demonstrate competitive or superior performance against both open-source and proprietary systems.
We release code, models, and the benchmark suite to support future research.
\end{abstract}

\begin{figure}[b]
  \centering
  \includegraphics[width=1\textwidth]{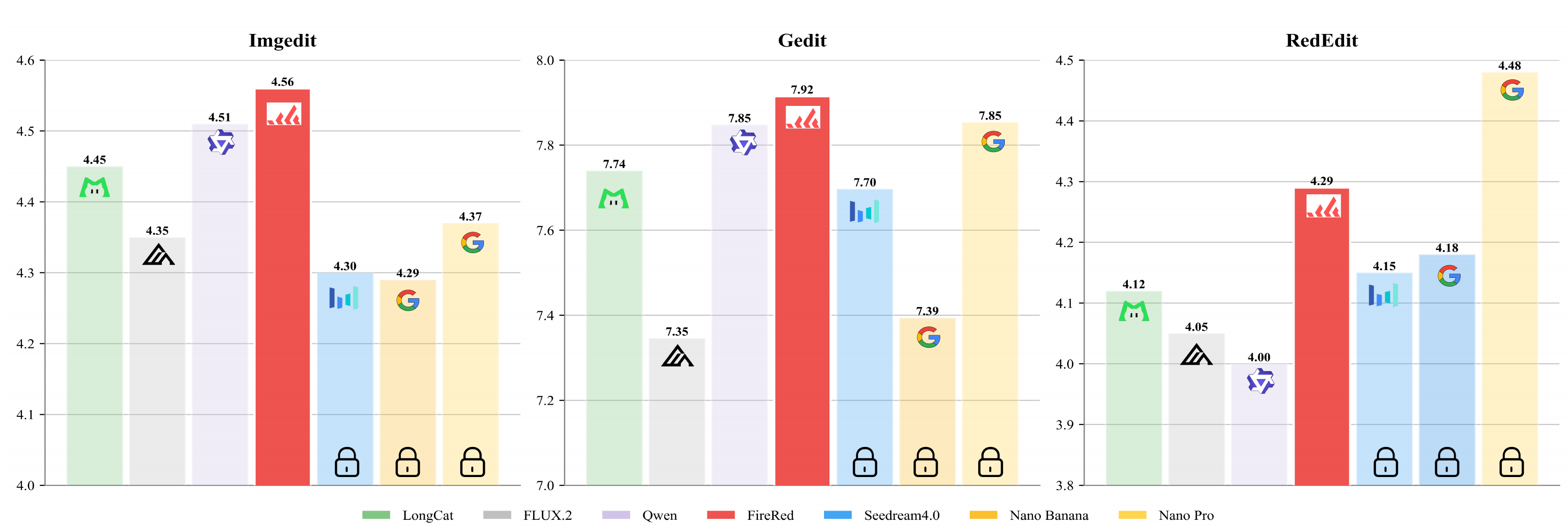}
  \caption{This figure benchmarks generative image models across four human evaluation dimensions (alignment, consistency, realism, aesthetics) and three editing tasks (Imgedit, Gedit, RedEdit). }
  \label{fig:introduction_benchmark}
\end{figure}

\begin{figure}
  \centering
  \includegraphics[width=1\textwidth]{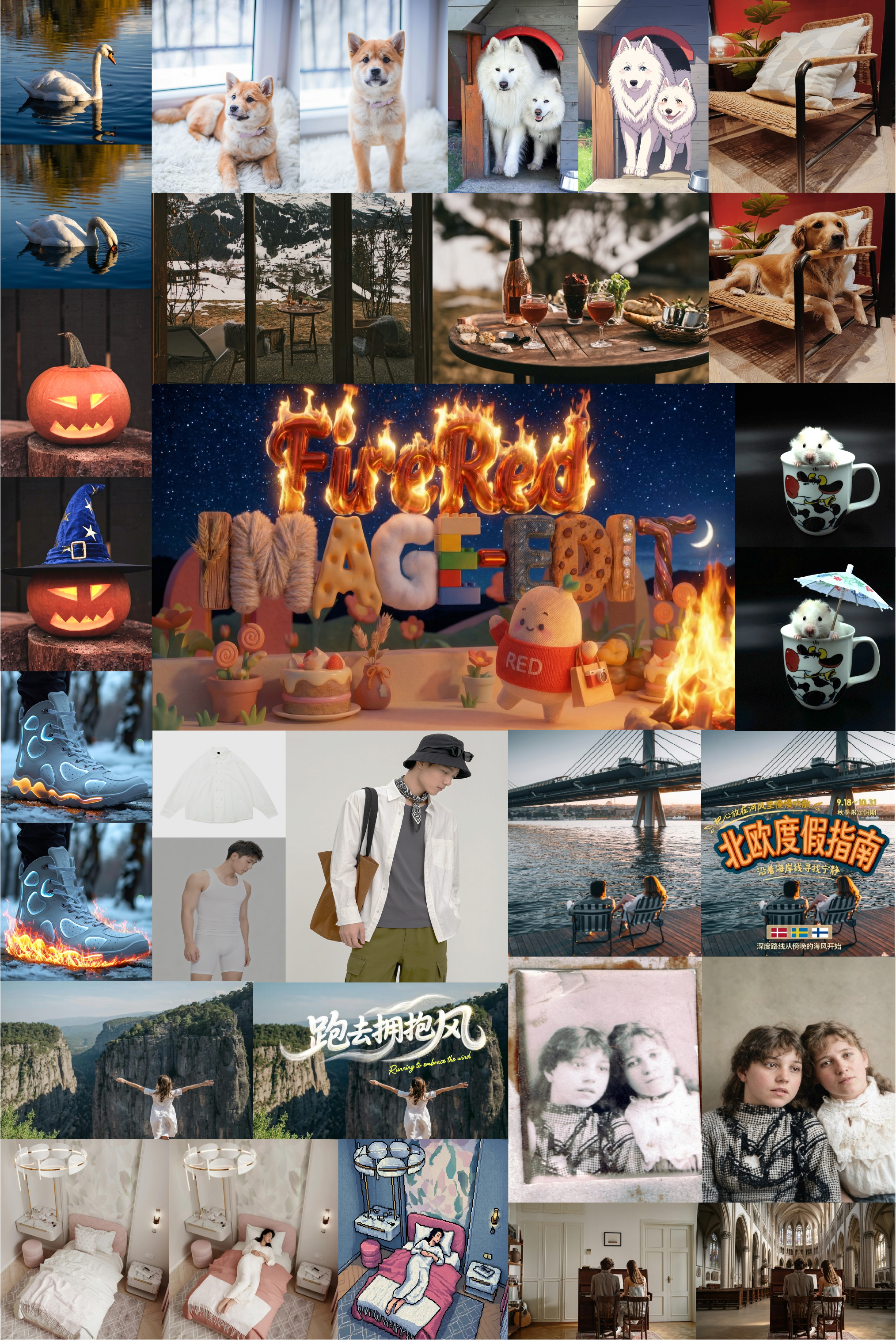}
  \caption{Showcase of \methodname in general image editing. }
  \label{fig:vis_introduction}
\end{figure}



\setlength{\parindent}{0pt}

\begin{tikzpicture}[remember picture,overlay]
            \node[anchor=north west, xshift=1.5cm, yshift=-1.0cm] at (current page.north west) {
                \includegraphics[width=5.4cm]{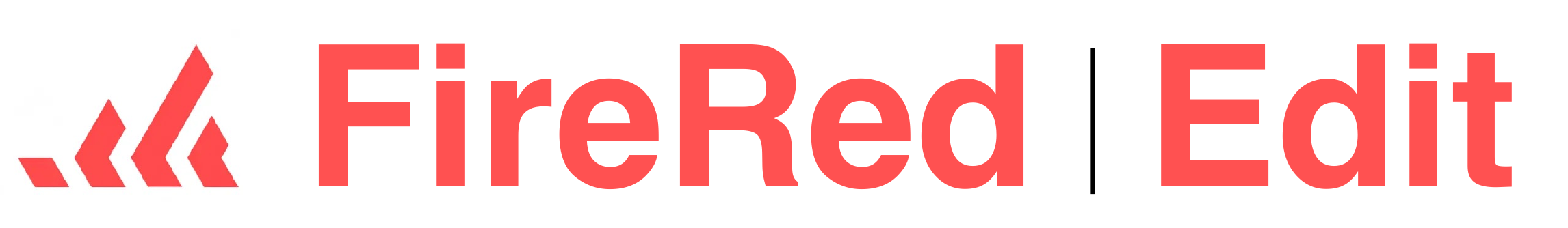}
            };
\end{tikzpicture}

\maketitle

\begin{center}
    \renewcommand{\arraystretch}{1.5}
    \begin{tabular}{rll}

        \github{} & \textbf{GitHub} & {\textcolor{sectionred}{\href{https://github.com/FireRedTeam/FireRed-Image-Edit}{https://github.com/FireRedTeam/FireRed-Image-Edit}}} \\
        \huggingface{} & \textbf{HuggingFace Model} & {\textcolor{sectionred}{\href{https://huggingface.co/FireRedTeam/FireRed-Image-Edit-1.0}{https://huggingface.co/FireRedTeam/FireRed-Image-Edit-1.0}}} \\
        \huggingface{} & \textbf{HuggingFace Demo} & {\textcolor{sectionred}{\href{https://huggingface.co/spaces/FireRedTeam/FireRed-Image-Edit-1.0}{Online Demo (HuggingFace)}}} \\
    \end{tabular}
\end{center}

\setcounter{footnote}{0}  
\renewcommand{\thefootnote}{\arabic{footnote}}  

\clearpage
\renewcommand{\contentsname}{\textcolor{sectionred}{Contents}}
\tableofcontents
\clearpage


\section{Introduction}
\label{sec:introduction}

The domain of text-to-image (T2I) synthesis has undergone a paradigm shift in recent years, progressing rapidly from elementary texture generation to the creation of photorealistic imagery with intricate semantic alignment~\cite{podell2023sdxl,esser2024scaling,flux2023,qwenimage,seedream2025seedream,liu2025step1x,betker2023improving}. However, this surge in performance has erected substantial barriers to entry. The current ecosystem is increasingly polarized: on one hand, proprietary commercial heavyweights—such as Nano Banana Pro~\cite{nanopro} and Seedream 4.0~\cite{seedream2025seedream}—operate as opaque "black boxes," delivering superior fidelity but offering zero interpretability or reproducibility. On the other hand, while the open-source community strives for democratization, the prevailing trend has been to relentlessly scale model size—ballooning to tens of billions of parameters (e.g., Qwen-Image~\cite{qwenimage} at 20B, FLUX.2~\cite{flux-2-2025} at 32B, and Step-1X-Edit~\cite{liu2025step1x} at 19B). This trajectory imposes unsustainable computational burdens for both training and deployment. Consequently, distilling synthetic data from proprietary models has become a popular workaround for resource-limited academic fields~\cite{lu2025generate,lai2025anomalypainter}. Despite these efforts, the community continues to fixate on the sheer scaling of model parameters, often at the expense of efficiency in data processing and model training. Critical gaps remain in two key areas: the efficient construction of high-quality datasets for image editing, and the development of a scientifically rigorous, standardized benchmark capable of evaluating these models across universal metrics.

In this paper, we introduce FireRed-Image-Edit, a holistic, end-to-end framework that rigorously optimizes every facet of the model's lifecycle-spanning data curation, architecture design, training efficiency, and inference optimization-enabling state-of-the-art performance with massive high-quality data.


To fully exploit this architecture, we scale our training regime to approximately 100 million diverse image-text pairs spanning text-to-image generation, single and multi-image-to-image synthesis, and complex instruction-based editing. Supporting this massive data scale demands systematic training efficiency innovations. We introduce a Multi-Condition Aware Bucket Sampler that explicitly accounts for variable input image counts across different editing tasks, optimizing aspect ratio batching to minimize padding-induced computational waste while preserving spatial layout integrity. Complementing this, our Stochastic Instruction Alignment mechanism employs random dropout and permutation of reference images during data collation, with dynamic re-indexing of text prompts to maintain semantic correspondence—forcing the model to decouple spatial orderings from content and thereby enhancing generalization in complex multi-reference scenarios. At the system level, we integrate pre-extracted text feature caching, Fully Sharded Data Parallel (FSDP) for distributed state management, gradient checkpointing, and mixed-precision training to maximize throughput without sacrificing stability.

Our training strategy further refines convergence through three coordinated mechanisms. Distributed Stratified Timestep Sampling decomposes the diffusion horizon into sub-intervals assigned across GPU ranks, with synchronized rotation preventing rank-specific overfitting and ensuring global uniformity in noise coverage. Logit-Normal Loss Weighting concentrates gradient contributions on the semantically critical intermediate diffusion timesteps while suppressing negligible extremes. Finally, Model Weight Averaging via Exponential Moving Average synthesizes parameters across the optimization trajectory, neutralizing transient fluctuations for enhanced robustness against real-world distribution shifts.

Furthermore, we recognize that conventional metrics often fail to capture the nuance required for production-ready applications, leading to a disconnect between academic scores and actual user experience. To bridge this gap, we establish a comprehensive, multi-dimensional benchmark designed to rigorously evaluate model performance closer to practical deployment standards. This evaluation framework encompasses a wide spectrum of editing tasks, ranging from subtle textural adjustments to complex structural alterations, and assesses the model across multiple axes—including instruction alignment, background preservation, and photographic fidelity. By shifting the evaluation focus from theoretical metrics to applied utility, our benchmark provides a granular analysis of current limitations and demonstrates our model's superior adaptability in facilitating genuine, unrestricted creative workflows.

\section{Data}
\label{sec:data-infrastructure}
The quality of training data is fundamental to generative models and largely sets their achievable performance. 
To this end, we collected 1.6 billion samples in total, comprising 900 million text-to-image pairs and 700 million image editing pairs. 
The editing data is drawn from diverse sources, including open-source datasets (e.g., OmniEdit~\cite{wei2024omniedit}, UnicEdit-10M~\cite{ye2025unicedit}), our data production engine, video sequences, and the internet, while the text-to-image samples are incorporated to preserve generative priors and ensure training stability. 
Through rigorous cleaning, fine-grained stratification, and comprehensive labeling, and with a two-stage filtering pipeline (pre-filter and post-filter), we retain 100M+ high-quality samples for training, evenly split between text-to-image and image editing data, ensuring broad semantic coverage and high data fidelity.

\subsection{Data Distribution}
\begin{figure}[H]
    \centering
    \includegraphics[width=\textwidth]{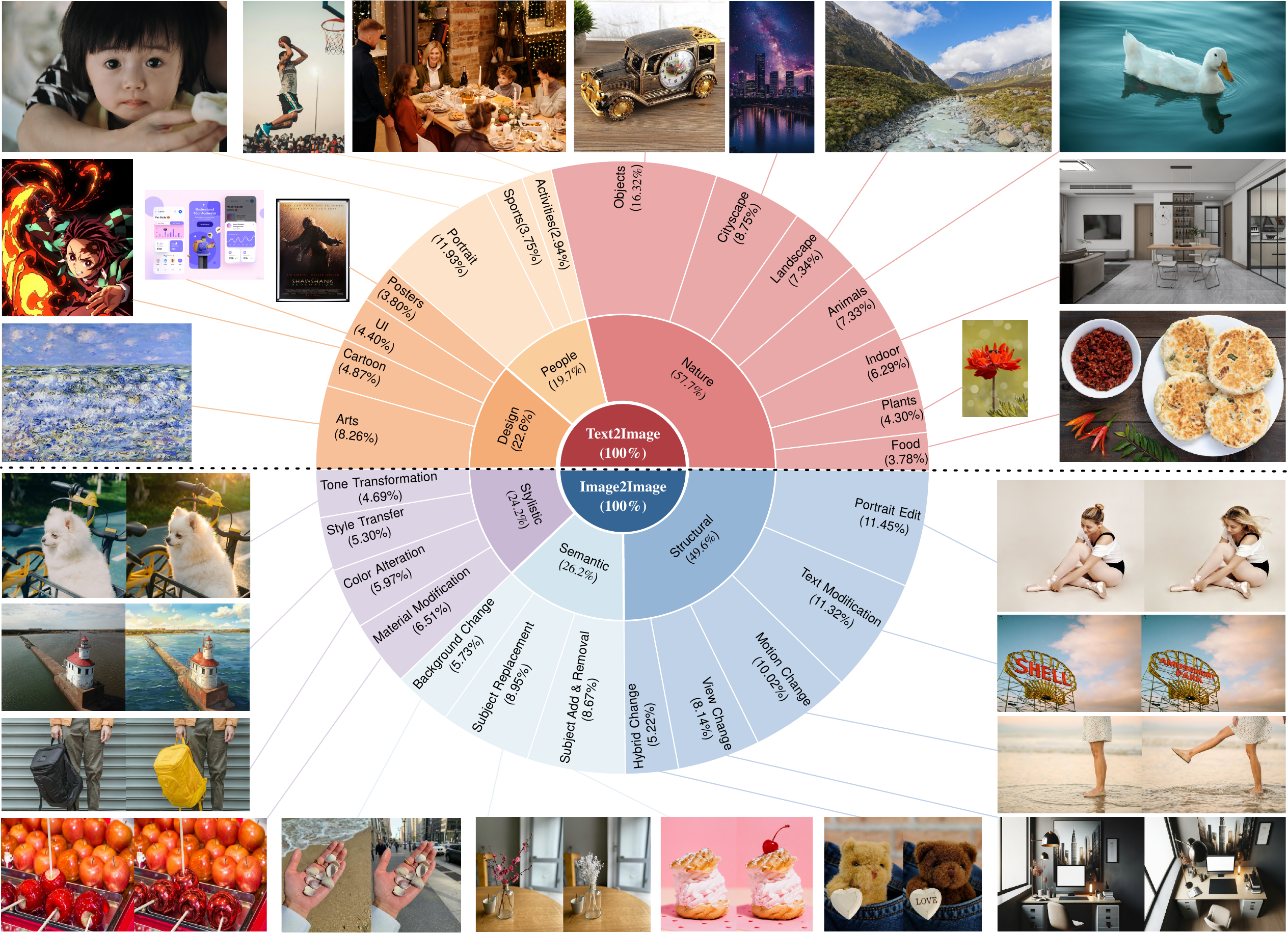}
    \caption{\textbf{Overview of Data Distrubution.} Our training dataset achieves an approximate 1:1 ratio between Text-to-Image (T2I) and Image-to-Image (I2I) tasks after data cleaning. The T2I component is divided into three main categories: Nature (general-purpose generation), People (human-centric generation), and Design (artistic styles, text rendering, and complex layouts). The I2I component is also divided into three categories: Semantic Editing (content-based modifications), Stylistic Editing (aesthetic adjustments), and Structural Editing (spatial arrangement and composition).
Our data collection strategy ensures a balance of diversity and quality throughout the training process, providing comprehensive coverage and precise annotations to foster robust model training.}
    \label{fig:pretrain_data_distribution}
\end{figure}

Our dataset is carefully curated to support the development of a robust image editing model. The dataset is divided into two main categories: Text-to-Image (T2I) and Image-to-Image (I2I). The data is meticulously balanced across various subcategories to ensure diversity and comprehensive coverage of real-world scenarios, as shown in Fig.~\ref{fig:pretrain_data_distribution}. After thorough filtering and selection, the dataset is ultimately approximately balanced in a 1:1 ratio between T2I and I2I data. This balanced composition is intentional, as the inclusion of T2I data is crucial for building a strong foundation in text semantic understanding, which enables the model to achieve better generalization on a wide variety of image editing tasks.

\paragraph{Text-to-Image (T2I)}
The T2I data is incorporated into the training process to preserve generative priors and ensure training stability, allowing the model to produce coherent and realistic images from textual prompts.
The dataset is divided into several key categories, each selected to cover a wide range of contexts, styles, and subject matter. The major categories in the T2I dataset are as follows:

\begin{itemize}
    \item \textbf{Nature (57.7\%)}: The largest category, covering diverse subcategories such as Objects, Landscapes, Cityscapes, Plants, Animals, Food, and Indoor scenes. This category is essential for generating realistic and varied natural environments.
    
    \item \textbf{Design (22.6\%)}: Focuses on structured and artistic visual content, including Posters, User Interfaces, Cartoon and other forms of art. It plays a key role in training the model to understand complex layouts, artistic styles, and detailed textual prompts.

    \item \textbf{People (19.7\%)}: Includes images related to human figures, such as Portraits, Sports, and Activities. This category is crucial for training the model to generate accurate and diverse human-related imagery, enhancing its ability to handle human-centered tasks.
\end{itemize}


\paragraph{Image-to-Image (I2I)}  
The I2I component of our dataset is crutial for enhancing or modifying existing images based on specific instructions. The tasks are broadly categorized into three main types: Semantic Editing, Stylistic Editing, and Structural Editing. These categories are defined by the nature of the modifications made to the image—whether it’s altering the content, enhancing the visual style, or adjusting the structure and composition of the scene. Below, we describe each category along with its associated subcategories:

\begin{itemize}
    \item \textbf{Semantic Editing (26.2\%)}: This category involves modifying the content of an image, such as changing the objects or background. It includes tasks like adding, removing, or replacing objects, and altering the background. These tasks are important for enabling the model to edit the primary elements of the scene, allowing for precise content manipulation while maintaining the overall coherence of the image.
 
    \item \textbf{Stylistic Editing (24.2\%)}: Stylistic editing focuses on enhancing the visual style and aesthetic characteristics of an image without altering its core content. Tasks in this category include color alteration, style transfer, tone transformation, and material modification. These tasks are crucial for training the model to change the image's appearance, such as modifying the atmosphere, applying artistic styles, or improving the visual aesthetics while preserving the subject matter.

    \item \textbf{Structural Editing (49.6\%)}: Structural editing covers changes to the spatial arrangement and positioning of elements within the image. It includes fine-grained, control-intensive scenarios such as view change, motion change, and portrait change. Beyond these, more complex cases like text modification and higher-level hybrid change are also included, requiring coordinated spatial and visual reasoning and enabling the model to perform more advanced and controllable scene transformations.
\end{itemize}


\subsection{Data Pre-Filtering}
\begin{figure}[!b]
    \centering
    \includegraphics[width=0.8\textwidth]{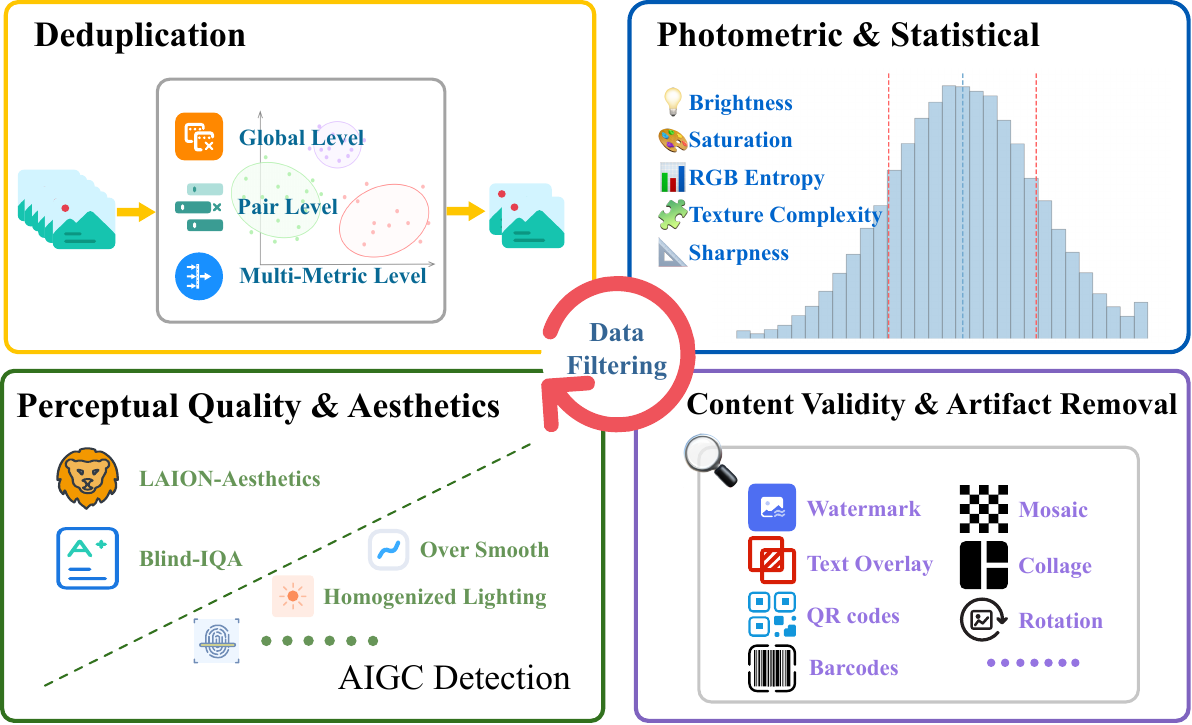}
    \caption{\textbf{Overview of Data Filtering.} Our multi-stage data filtering pipeline includes deduplication, photometric and statistical filtering, artifact removal, perceptual quality assessment, and AIGC detection. These steps ensure the dataset is free from redundancy, visual noise, artifacts, and synthetic content, maintaining high-quality samples for image editing model training.
}
    \label{fig:data_filter}
\end{figure}

\subsubsection{Deduplication}
To mitigate redundancy from diverse sources and ensure high-quality training data, we implement a multi-level hierarchical deduplication pipeline, which consists of three levels with increasing strictness:

\textbf{Level 1: Coarse Global Deduplication}
At the first level, we perform a large-scale near-duplicate retrieval and clustering across all images using a scalable image deduplication system~\cite{pizzi2022self}. Images are embedded into a shared feature space, where we identify and group visually similar samples, leaving only representative samples that contribute distinct visual content to the dataset. This stage primarily addresses exact duplicates and broad similarities across the image pool.

\textbf{Level 2: Pair-Level Deduplication}
The second level focuses on image-to-image (I2I) editing data, where redundancy can also manifest in the form of duplicate source–target pairs. In this stage, we apply a coarse pair-level deduplication, where CLIP~\cite{radford2021learning} embeddings and an internal image embedding model are used to compute similarity scores between source and target images. Pairs with high semantic similarity are discarded to remove trivial identity mappings or identical edit results. This helps in reducing template-based redundancy that can negatively affect model generalization.

\textbf{Level 3: Fine-Grained Multi-Metric Deduplication}
At the final level, we apply a strict multi-metric deduplication strategy to ensure the retention of only genuinely distinct and meaningful edit samples. Here, we combine Peak Signal-to-Noise Ratio (PSNR)~\cite{gonzalez2018digital}, Structural Similarity Index Measure (SSIM)~\cite{wang2004image}, Image Similarity Challenge 2021 metric (ISC21)~\cite{douze20212021}, and CLIP similarity to perform a fine-grained comparison. This multi-metric approach allows us to eliminate subtle near-duplicates, templates with minimal variation, and images that exhibit visually insignificant changes. By applying this stringent filtering, we ensure that the dataset retains only high-quality, diverse, and non-repetitive edit samples.


\subsubsection{Photometric \& Statistical Filtering}
To reduce visual noise and stabilize training, we perform photometric and statistical filtering using descriptors like Brightness, Saturation, RGB Entropy, Texture Complexity, and Sharpness. Brightness and saturation are used to detect under- or over-exposed images and those with abnormal color responses due to extreme lighting or non-natural rendering. RGB Entropy filters out images with large uniform regions or flat backgrounds, while Texture Complexity removes samples with irregular high-frequency patterns, often caused by compression artifacts or non-semantic textures. Sharpness is used to identify images with motion blur, defocus, or other degradations that obscure fine details. We estimate the empirical distribution of each descriptor and discard samples outside the statistically plausible range, ensuring the dataset retains stable photometric properties and richer structural content for training.



\subsubsection{Content Validity \& Artifact Removal}
To ensure valid content and prevent misleading supervision signals, we use specialized detectors to remove images with undesirable artifacts. This includes filtering images with watermarks, visible text overlays, mosaics, collages, privacy masks, QR codes, barcodes, as these can introduce non-semantic patterns or distortions that impair model performance. We also apply rotation correction based on metadata and geometric cues to ensure consistent spatial alignment, reducing unnecessary geometric variance. This filtering improves the validity of visual supervision and provides a cleaner foundation for accurate image editing.




\subsubsection{Perceptual Quality \& Aesthetics}
We use deep learning evaluators to assess image quality and aesthetics. IQA metrics~\cite{mittal2012making,mittal2012no} detect issues like blur, noise, and exposure imbalance, while aesthetics predictors~\cite{schuhmann2022laion} evaluate composition and appeal. Low-quality or poorly rated samples are filtered out, ensuring better visual coherence for image editing training.

\subsubsection{AIGC Detection}
We use internal AIGC detection to identify and filter images generated by generative models. These samples may bypass standard quality checks but exhibit biases like smooth textures, uniform lighting, and suppressed details. Removing AIGC content preserves authentic image statistics and prevents synthetic priors from degrading model generalization.

\subsection{Data Production Engine}
\begin{figure}[h]
    \centering
    \includegraphics[width=\textwidth]{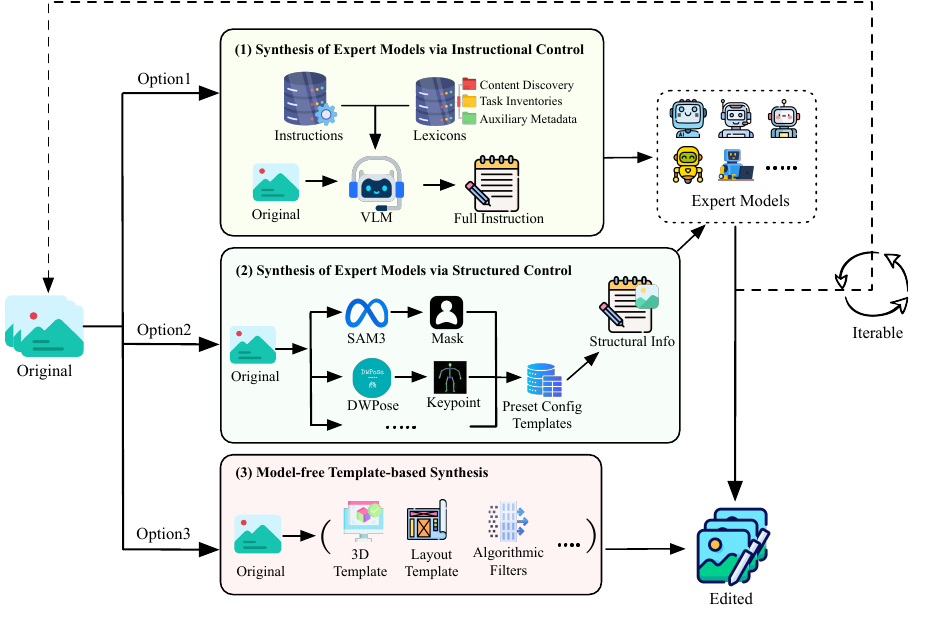}
    \caption{\textbf{Overview of Data Engine.} The data production engine generates paired image editing samples through three forward construction strategies. (1) Instructional Control synthesizes expert models using instruction templates and edit-target lexicons grounded by VLM discovery and auxiliary metadata. (2) Structured Control leverages structural priors such as masks and pose keypoints extracted from perception modules to guide expert models with precise control signals. (3) Model-free Template-based Synthesis includes approaches such as predefined 3D templates, layout templates, and algorithmic filters to enable controllable and deterministic generation. The pipeline is designed to be iterable, supporting complex multi-step edits.}
    \label{fig:data_engine}
\end{figure}

While real-world data offers a natural distribution, it often lacks the density required to support specialized editing tasks. 
To address this limitation, we build a data production pipeline that generates paired editing data through three forward construction strategies, as illustrated in Fig. \ref{fig:data_engine}, enabling dense coverage and controllable synthesis across the editing task space.

\paragraph{Synthesis of Expert Models via Instructional Control}
We adopt image editing models~\cite{zhang2023adding,wu2025qwenimagetechnicalreport,LongCat-Image,labs2025flux1kontextflowmatching} that require only the input image and an instruction as the default experts, enabling fast and broad data generation. Building on this foundation, we develop an instruction-driven expert synthesis pipeline to stably map diverse editing tasks to the capability boundaries of different expert models. 
Concretely, we formalize each expert’s scope into deterministic instruction templates and build and maintain: an Expert-instruction library and an Edit-target lexicon. The Expert-instruction library summarizes model constraints and triggering conditions into canonical primitives and structured slots, improving instruction stability and performance. The Edit-target lexicon defines what to edit by combining VLM Discovery to interpret the image and propose grounded edit targets, Task Inventories as a task-aligned library of editable objects/attributes for long-tail coverage, and auxiliary metadata (e.g., OCR, coordinates) as computable anchors for precise grounding. To further improve editing quality, we refine expert models with SFT or LoRA on curated data.


\paragraph{Synthesis of Expert Models via Structured Control
}
We further introduce expert models~\cite{zhang2023adding,wang2025dynamicface,xiang2025instanceassemble,xu2025single,liao2024appearance,zhou2024storymaker,he2025flux,wang2025target,guo2024liveportrait} that require structured control, where editing is driven by non-text, structured signals to reduce language ambiguity and improve generation controllability. 
Specifically, we extract structural priors from auxiliary perception modules such as SAM~\cite{carion2025sam} and DWpose~\cite{yang2023effective}, including segmentation masks and pose keypoints, and combine them with task parameters in Preset Configuration Templates to derive task-specific structured control info. 
The structured info is fed into expert models to precisely modulate the editing process.
This structured-control approach is particularly effective for spatially sensitive tasks such as precise object removal, facial expression and body pose transfer, and appearance redirection.

\paragraph{Model-free Template-based Synthesis} 
To complement generative approaches, we adopt a model-free template-based synthesis pipeline that avoids model inference, instead utilizing a rich library of predefined algorithmic templates to produce editing pairs with precise edits and without artifacts. 
The template set includes, but is not limited to: 3D Parametric Templates that leverage graphics engines and skeletal rigs to synthesize pixel-consistent facial expressions, body re-posing, and other parametric transformations; Structured Layout Templates that define spatial anchors for text, logos, UI elements, and similar graphic components to strictly enforce typographic and layout constraints; and Algorithmic Filter Templates that cover deterministic image signal processing operations such as sharpening, color redirection, and other low-level visual enhancements. Beyond these, we incorporate additional task-specific templates to support a wider range of deterministic edits.

Beyond the three forward construction strategies, we further enhance data fidelity and complexity through two mechanisms: Task Inversion and Task Splitting. Task Inversion leverages the reversibility of certain edits by swapping source and target images, thereby reducing instruction direction bias. Task Splitting addresses high-difficulty requests by decomposing multi-operation edits into sequential atomic steps, applying our iterable pipeline that performs one modification at a time, thereby reducing quality degradation and preserving structural consistency.

\subsection{Long-tail Data Supplementation.}
To deal with the long-tail problem where some editing tasks lack enough samples, we use a straightforward "check and fill" strategy.

We first evaluate the coverage of our existing dataset by indexing all training instructions into a vector database. When the model exhibits poor performance in a specific editing domain, we query this index to quantify the available training samples. If the sample density for a particular instruction is insufficient, we identify it as a "data gap" that requires targeted supplementation.

Once a gap is identified, we leverage our robust image retrieval framework to acquire base imagery from a massive candidate pool. 
This framework employs a specialized deep feature encoder to map raw imagery into a high-dimensional latent space, generating dense, highly discriminative embeddings. Supported by a large-scale vector indexing engine, the system utilizes seed images or semantic queries to retrieve samples exhibiting high fidelity in spatial layout, texture, color distribution, and semantic coherence via metric learning. Using both text-to-image and image-to-image retrieval, we can precisely locate images that match the missing task's requirements. These retrieved images are then processed through our data production engine to generate new, high-quality training pairs, effectively rectifying the weak points in our data distribution.

\subsection{Captioning Engine}
To provide high-quality textual supervision for both Text-to-Image (T2I) generation and Image-to-Image (I2I) editing, we developed a comprehensive Captioning Engine. This engine operates progressively: from static single-image understanding to dynamic cross-image differential reasoning, and finally to user-centric instruction refinement.
\subsubsection{Structured Captioning}
For both T2I training and the visual grounding of I2I pairs, simple alt-texts are insufficient. We employ a VLM to generate Structured Captions (see Annotation Prompt~\ref{pb:Annotation Prompt}) that provide a comprehensive breakdown of the image. These captions explicitly detail the main subject, background elements, lighting conditions, artistic style, and camera angles. This "structured" supervision ensures the model understands what is in the image before learning how to change it.

\begin{figure}[h]
    \centering
    \includegraphics[width=\textwidth]{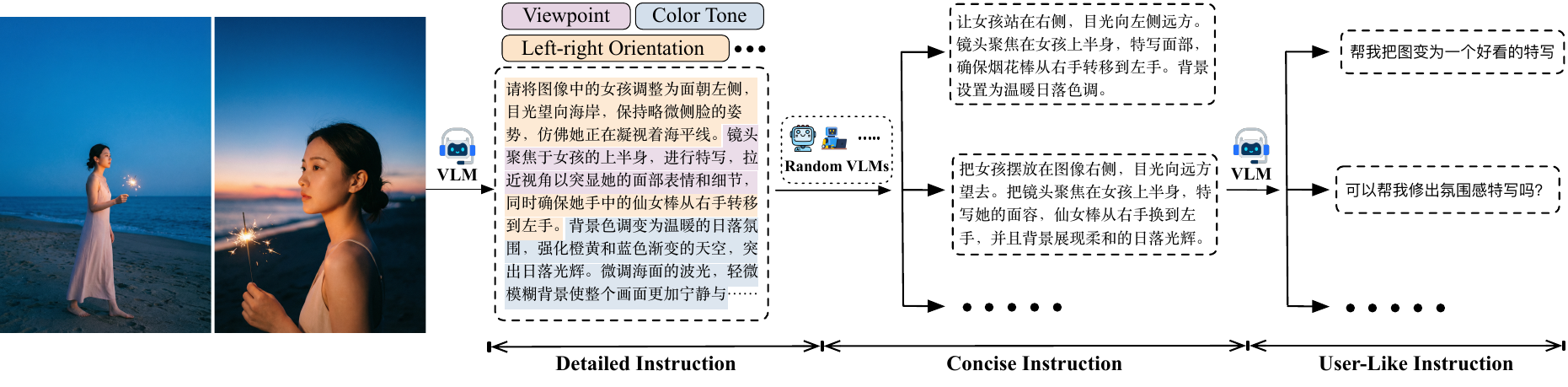}
    \caption{\textbf{An example of Instruction Captioning}. Detailed Instruction Captioning generates precise descriptions of visual changes, Concise Instruction Refinement simplifies these instructions while retaining key details, and User-Like Instruction Refinement rephrases them in a more natural, conversational tone.}
    \label{fig:data_caption}
\end{figure}

\subsubsection{Instruction Captioning}
To provide high-quality textual supervision for I2I (Image-to-Image) editing tasks, we categorize our instruction generation into three distinct types: Detailed Instruction Captioning, Concise Instruction Refinement, and User-Like Instruction Refinement.

\paragraph{Detailed Instruction Captioning}
Leveraging a specially fine-tuned VLM~\cite{Qwen3-VL}, this stage focuses on generating foundational ground truth benchmarks for visual transitions between image pairs. The objective is to produce descriptions that are accurate, complete, and unambiguously reflective of the source-to-target changes. The model is specifically trained to conduct high-fidelity differential reasoning, consistently modeling subject attributes, spatial relationships, and semantic shifts. Special emphasis is placed on high-sensitivity dimensions, such as left-right orientation, viewpoint variations, and fine-grained attributes, which demand superior recognition and expressive capabilities from the model.

\paragraph{Concise Instruction Refinement}
This stage concurrently addresses both instruction brevity and linguistic diversity, integrating them into a unified refinement pipeline. By utilizing random VLM models from a suite of diverse VLM models~\cite{glm2024chatglm,team2025kimi} and extensive lexical libraries, we implement varying levels of simplification while strictly preserving the original editing intent. To prevent the model from overfitting to the rigid templates of the base Caption VLM, we dynamically sample diverse syntactic structures during the rewriting process, ranging from inversions and subjectless imperatives to informal phrasings and various other linguistic patterns. This approach facilitates intent purification by stripping redundant visual modifiers and abstracting concrete concepts, ensuring the model masters core editing logic rather than specific object-based patterns.

\paragraph{User-Like Instruction Refinement}
This stage refines the concise outputs into natural, everyday language to better reflect how people actually speak. The goal is to bridge the gap between technical commands and human conversation by injecting persona-driven tones and help-seeking markers. This simulates real-world scenarios where a user might request help using phrases such as "Can you help me fix this?" or "Could you save this image by changing something?". By incorporating these warm and interactive phrases at varying ratios, we significantly enhance the model’s ability to interpret informal, intuitive, and help-seeking user intents.

By mixing these rewritten concise instructions with the original detailed descriptions for training, we constructed a multi-scale data distribution covering the spectrum from minimal colloquialisms to fine-grained specifications. This mixed strategy allows the Fine-grained Visual Reasoning capabilities learned from long texts to be effectively generalized to short text scenarios. Consequently, the model can leverage its deep visual knowledge to perform accurate inference and execution, even when facing ambiguous or concise user instructions.


\subsection{Data Post-Filtering}
\label{subsubsec:data-cleaning}
To ensure high fidelity and precise semantic alignment, we implement a post-filtering pipeline for automated data quality assessment and curation. This stage serves as an automated quality gate to refine data including both natural real-world pairs and synthesized pairs. Specifically, we train a specialized multimodal evaluation model based on Qwen3-VL-8B~\cite{Qwen3-VL} by integrating LLM-driven hard negative mining with expert-level human annotation.
This model precisely quantifies edited images across multiple dimensions. Subsequently, we leverage this model to perform automated filtering on a dataset of more than a hundred million raw samples. By eliminating instances exhibiting semantic misalignment or poor perceptual quality, we significantly enhance the overall distributional quality of the training dataset.


\paragraph{Hard Negative Mining and Multi-Dimensional Annotation} High-quality negative samples are critical for enhancing the discriminative capability of evaluation models during data refinement. We initially sample 50,000 triplets of "Source Image – Editing Instruction – Target Image" from the massive raw dataset to serve as positive samples. Subsequently, we employ an LLM to apply semantic perturbations and stochastic rewriting to the original instructions. This process generates deviate instructions to construct 50,000 negative samples, resulting in a balanced dataset of 100,000 instances.
To establish a gold-standard benchmark, we organize a group of human experts for double-blind annotation. The evaluation criteria focus on two key metrics: instruction alignment (assessing whether the edit accurately executes the textual command) and perceptual quality (checking clarity, artifacts, and aesthetic appeal). This rigorous annotation scheme ensures a balanced semantic distribution, providing a robust foundation for the precision of the subsequent evaluation model.

\paragraph{Training of the Vision-Language Evaluation Model} To achieve an automated assessment mechanism that aligns closely with human aesthetic and logical judgments, we utilize Qwen3-VL-8B as the backbone for fine-tuning. Through supervised fine-tuning (SFT) on the aforementioned 100,000 high-quality annotated samples, the model acquires the capability of parsing complex editing instructions and perceiving subtle visual discrepancies.
Experimental results demonstrate that the evaluation model exhibits exceptional robustness in multi-dimensional scoring tasks. Its metrics show a strong correlation with the evaluations of human experts. Beyond identifying subtle semantic deviations, the model is highly sensitive to structural distortions and textural artifacts in generated images, serving as a reliable automated proxy to replace costly manual evaluation.

\paragraph{Large-Scale Data Curation and Quality Assurance} Using the trained evaluation model, we conduct a comprehensive quality assessment of the raw training data at a scale of over 100 million samples. We establish strict filtering thresholds to perform joint screening based on two dimensions: alignment score and quality score.
Specifically, the model automatically identifies and discards samples that fail to follow the instructions, exhibit semantic shifts, or possess low visual fidelity (e.g., blurring, artifacts, or logical inconsistencies). This systematic data purification process significantly refines the distribution of the training set. It ensures that the generative model focuses on high-quality, high-fidelity editing pairs during subsequent large-scale training, thereby fundamentally improving the performance of the final model.

\section{Model Training}
\label{sec:model-training}

\begin{figure}[h!]
  \centering
  \includegraphics[width=1\textwidth]{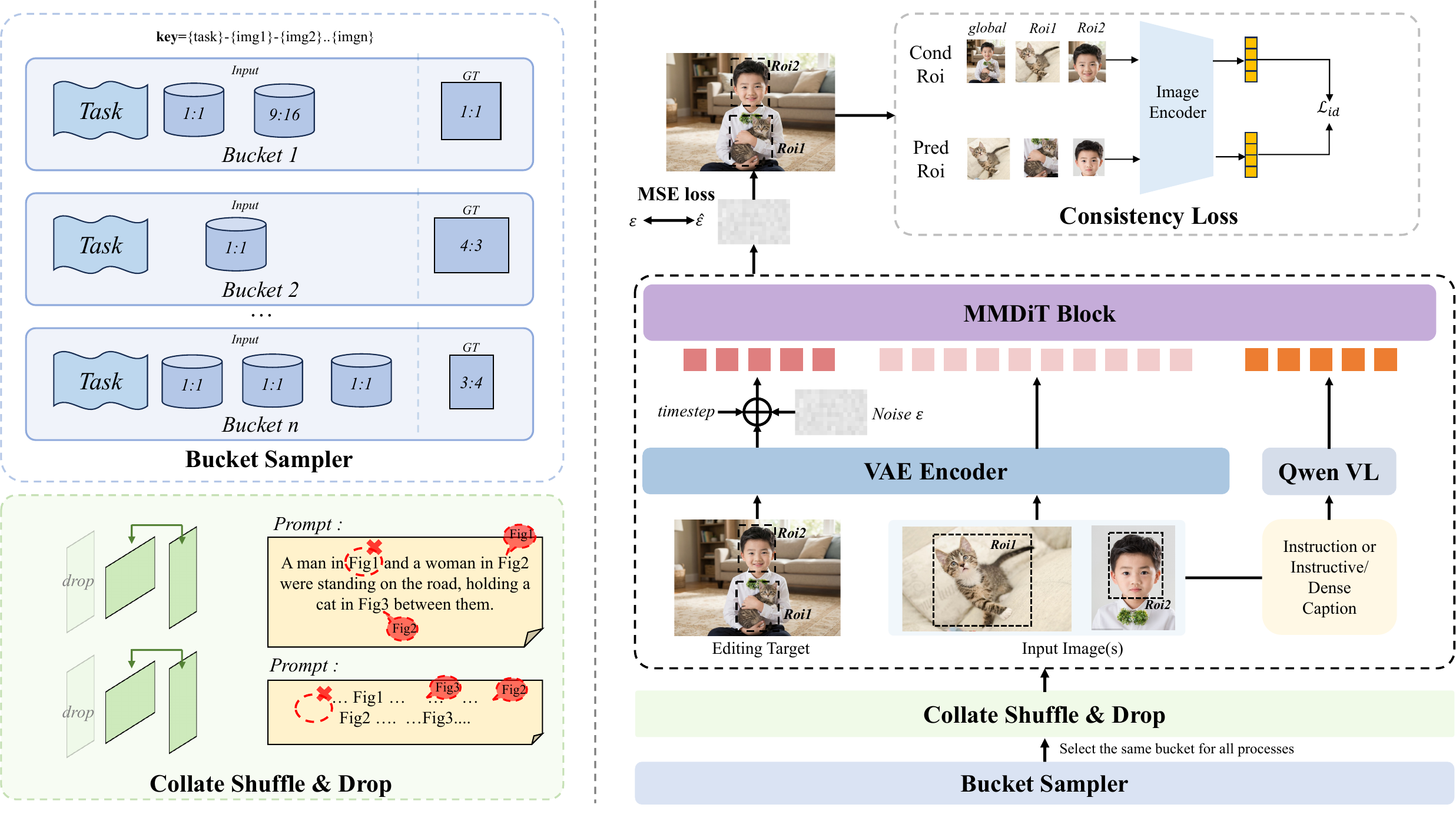}
  \caption{\textbf{Architecture overview}. The data pipeline begins with a Bucket Sampler that organizes input sequences based on task categories and aspect ratios to handle variable resolutions efficiently. This is followed by a Collate Shuffle \& Drop mechanism that augments text prompts by randomly permuting or omitting figure identifiers to enhance robustness. The core model employs MMDiT Blocks to process multimodal features, where visual inputs are encoded by a VAE Encoder and multimodal conditions (reference images and textual instructions) are processed by Qwen VL. To ensure high-fidelity generation, the training process incorporates an consistency similarity loss; this mechanism extracts regions of interest (RoIs) from both predicted and ground truth images, passing them through a shared image encoder to minimize identity discrepancy.}
  \label{fig:model_arch}
\end{figure}

\subsection{Architecture}
\label{sec:architecture}
Scalability and instructional fidelity are the cornerstone objectives guiding the design of our proposed model. Built upon an open-source multimodal text-to-image foundation~\cite{qwenimage}, our architecture inherits a profound understanding of vision-language nuances, which we further extend to the generative and editing domains. To bridge the gap from multimodal perception to high-fidelity synthesis, we scale our training regime using a curated dataset of approximately one billion image-text pairs. This extensive corpus encompasses a diverse array of tasks, including text-to-image, single/multi-image-to-image generation, and complex instruction-based editing. Such a massive data scale enables the model to reconcile abstract editing commands with pixel-level semantic transformations, ensuring robust generalization across diverse domains.

Following the paradigm of unified sequence modeling, we employ a Double-Stream Multi-Modal Diffusion Transformer (MM-DiT) as the core generative engine. In this configuration, text embeddings derived from the ~\cite{qwenimage} backbone, latent tokens from a high-fidelity VAE, and—in editing scenarios—reference image features are concatenated into a unified input stream. This setup facilitates dense bidirectional interactions between modalities, maximizing the model’s ability to preserve structural integrity while executing precise modifications. To manage the high-dimensional sequences, we utilize 3D Unified RoPE, wherein reference and target image tokens share spatial coordinates but are differentiated by a temporal interval. This coordinate alignment, coupled with distinct time-conditioning for clean reference and noisy target latents, empowers our model to achieve state-of-the-art (SOTA) performance across multiple competitive benchmarks, demonstrating exceptional stability and stylistic consistency in complex image manipulations.

\subsection{Training Efficiency Optimization}
\label{sec:traning_efficiency_optimization}
\noindent \textbf{Multi-Condition Aware Aspect Ratio Batching.} To facilitate efficient distributed training across tasks ranging from Text-to-Image (T2I) generation to complex multi-image referenced editing, we designed a \textbf{Multi-Condition Aware Bucket Sampler}. Unlike traditional samplers that rely solely on single-image resolutions, our approach explicitly accounts for the variable number of input images $N$ inherent in different editing instructions. To minimize synchronization overhead and ensure balanced computational loads across GPUs, we define a bucket $\mathcal{B}_{r,n}$ characterized by a target aspect ratio $r$ and the count of input images $n$. For a given batch, the total visual sequence length $L_{vis}$ is strictly constrained to ensure alignment with the hardware's computational capacity, formulated as $L_{vis} = \sum_{i=1}^n \lceil \frac{H_i \cdot W_i}{p^2} \rceil \approx C$, where $p$ denotes the patch size and $C$ represents a constant token capacity per device. This strategy effectively reduces GPU idle time caused by uneven token lengths.

To preserve the structural integrity of diverse spatial layouts (e.g., 1:1, 9:16, 3:4) and ensure tensor consistency within a distributed batch, we minimize the cropping transformation $\mathcal{T}$ applied to a set of images $\{I_1, \dots, I_n\}$. We define the optimal bucket dimensions $(h, w)$ by minimizing the aggregate cropping area, ensuring that the information loss is mathematically minimized:
\begin{equation}
    \arg \min_{(h,w) \in \mathcal{S}} \sum_{i=1}^{n} |(H_i \cdot W_i) - (h \cdot w)|
\end{equation}
where $\mathcal{S}$ represents the set of predefined resolution buckets. This optimization ensures that all images within a synchronized batch across distributed nodes share identical tensor dimensions, significantly reducing padding-induced computational waste while maintaining high-fidelity aspect ratio consistency for various input scales.

\textbf{Stochastic Instruction Alignment.} To enhance the model's robustness against input permutations and prevent over-fitting to specific reference patterns, we implement a \textbf{Collate Shuffle \& Drop} mechanism. During the data collation phase, reference images are subjected to a random dropout probability and a stochastic permutation of their input order. Crucially, to maintain semantic alignment between visual tokens and linguistic references, the text prompt is dynamically updated to reflect these spatial changes. For instance, if an image originally tagged as ``Fig 1'' is permuted to the position of ``Fig 2'' due to shuffling, the corresponding tokens in the instruction prompt are re-indexed to ensure the text correctly references the shuffled image sequence. This unified augmentation strategy forces the model to decouple specific spatial orderings from semantic content, fostering superior generalization in complex, multi-reference editing scenarios without relying on rigid input sequences.

\textbf{System-Level Efficiency and Stability.} To address the substantial memory demands of the large-scale DiT architecture, we implemented a series of system-level optimizations to maximize training throughput. A key strategy involved decoupling the VLM encoding process by pre-computing embeddings and storing them as offline tensors. This eliminates redundant online computation and reduces per-iteration FLOPs, allowing the system's compute power to be dedicated entirely to the generative transformer.

Building upon the memory headroom gained from offline feature extraction, we managed distributed model states using Fully Sharded Data Parallel (FSDP). By sharding optimizer states and gradients, we transformed the memory complexity of these overheads from $O(M)$ to $O(M/N)$, where $N$ is the number of GPUs. To further minimize the peak memory footprint, we integrated gradient checkpointing and utilized BF16 mixed-precision training.

Crucially, the cumulative reduction in VRAM overhead allowed us to transition from a global FSDP approach to the more communication-efficient Hybrid Sharded Data Parallel (HSDP) scheme. While standard FSDP shards model states across the entire cluster—often leading to communication bottlenecks over inter-node links—HSDP constrains sharding to within individual nodes. This allowed us to fully exploit high-speed intra-node interconnects (e.g., NVLink) for scale-up efficiency, while utilizing traditional data parallelism across nodes. By reducing the volume of inter-node synchronization, we harmonized our scale-out strategy with the available network bandwidth, resulting in a marked increase in overall training throughput.

\begin{figure}[h!]
  \centering
  \includegraphics[width=0.8\textwidth]{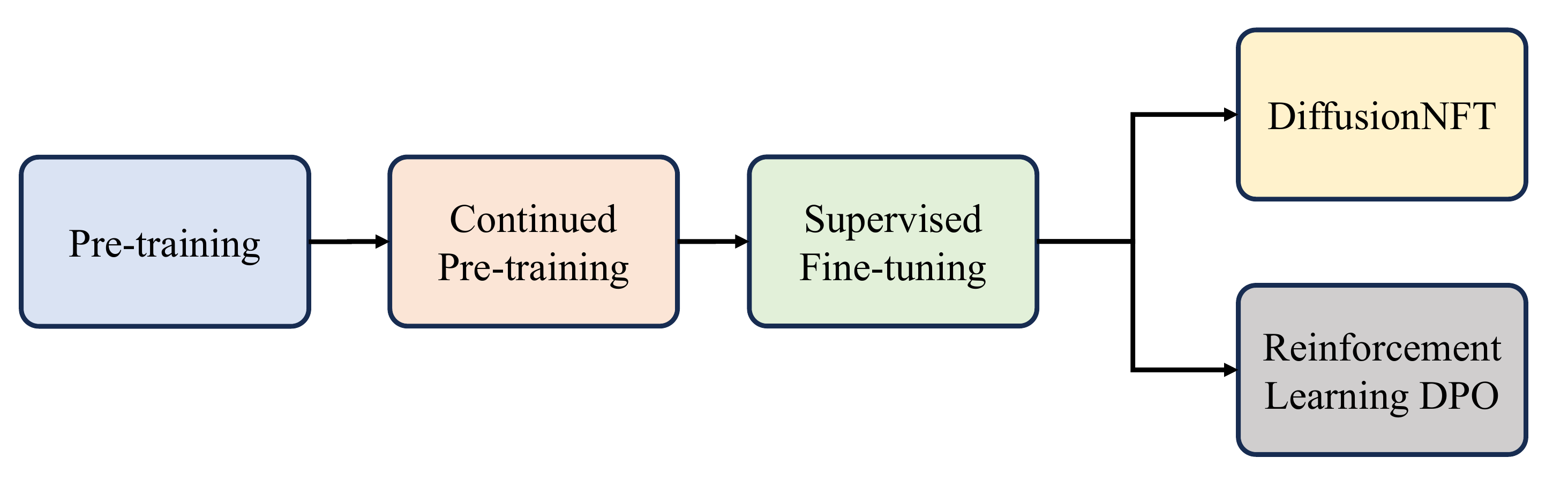}
  \caption{As shown in the figure, our training pipeline comprises progressive stages. We begin with Pre-training and Continued Pre-training to establish robust semantic and aesthetic foundations. The model then undergoes Supervised Fine-tuning to align with editing instructions, followed by Reinforcement Learning (DPO), culminating in the final DiffusionNFT model with enhanced generation quality and adherence.}
  \label{fig:train_pipeline}
\end{figure}

\subsection{Pre-training}
\label{sec:pretrain}
During the foundational pre-training phase, our primary objective is to establish a comprehensive visual vocabulary and a robust understanding of world knowledge. Unlike subsequent fine-tuning stages that prioritize aesthetic quality, this phase emphasizes data scale and semantic diversity. We adopt a holistic approach comprising three key pillars:

\textbf{Scalable Framework with Arbitrary-Resolution Adaptation.} To accommodate the diverse nature of real-world imagery, we implement a highly efficient and flexible training framework. Rather than adhering to a fixed, square resolution—which often necessitates aggressive cropping or resizing that corrupts image composition—we employ a dynamic bucket sampling strategy. This mechanism automatically groups images of similar aspect ratios into predefined buckets, allowing the model to train on arbitrary resolutions and aspect ratios (VAR) natively. This approach not only preserves the semantic integrity of the visual data but also optimizes computational efficiency by minimizing padding. Furthermore, our framework is architected to seamlessly alternate between text-to-image and single-image input modes. This flexibility enables the model to leverage both paired image-text data for semantic alignment and unpaired image data for purely visual feature learning, maximizing the utilization of available compute resources.

\textbf{Massive-Scale Data Integration with Hybrid Conditioning.} Our data strategy prioritizes the breadth of information, leveraging a massive corpus of data collected from the open internet. Acknowledging the inherent long-tail distribution and variable quality of web data, we adopt an inclusive filtering protocol. Instead of enforcing strict aesthetic curation or artificial class balancing—which might discard rare but semantically valuable concepts—we integrate all "effective" data samples. This allows the model to learn a vast, albeit noisy, representation of the visual world. To address the noise in web texts, we utilize a hybrid captioning scheme where the model is co-trained on: (1) \textit{Original Metadata}, preserving raw, noisy descriptions to capture rich, unstructured world knowledge and diverse linguistic styles; and (2) \textit{Structured Synthetic Captions}, generated to provide precise, tag-heavy descriptions that enhance the model’s prompt-following capabilities and object recognition. This mixture ensures the model acquires generalized knowledge while maintaining the ability to respond to structured prompts.

\textbf{Curriculum Learning via Progressive Timestep Sampling.} Given the high variance in data quality and the massive scale of the dataset, training stability is paramount. We introduce a progressive timestep curriculum to govern the noise diffusion process. In the initial phase, sampling is biased towards high-noise timesteps. Since high noise levels correspond to the low-frequency components of an image, this strategy forces the model to prioritize learning global semantic structures, layouts, and large-scale visual patterns, which are generally robust even in lower-quality data. As training converges, we gradually shift the distribution towards uniform sampling across all timesteps. This transition allows the model to refine its generation of high-frequency details and textures. This curriculum effectively mitigates the risk of divergence when training on noisy internet data and ensures a smooth transfer from learning coarse concepts to mastering fine-grained details.

\begin{table}[htbp]
\centering
\caption{Training Stages and Hyperparameters}
\label{tab:training_details}
\resizebox{0.75\textwidth}{!}{%
\begin{tabular}{llccccc}
\toprule
\textbf{Stage} & \textbf{Attribute} & \textbf{Pretrain} & \textbf{CT} & \textbf{SFT} & \textbf{DPO} & \textbf{NFT} \\
\midrule
\multirow{2}{*}{Dataset} 
 & Internet Collection & 100M & 5M & 50K  & 10K & 10K \\
 & Synthetic data      & 5M & 10M & 50K & 50K & 10K \\
\midrule
\multirow{3}{*}{Captioning} 
 & Original Cap. & 55\% & 20\% & 10\% & 10\% & 10\% \\
 & Structured Cap. & 40\% & 60\% & 45\% & 30\% & 30\% \\
 & Instructive Cap. & 5\% & 20\% & 45\% & 60\% & 60\% \\
\midrule
\multirow{6}{*}{Train Recipe} 
 & Resolution(pixel) & 384-512 & 512-1024 & 1024 & 1024 & 1024 \\
 & Training Steps & 300K & 65K & 5K & 5K & 500 \\
 & Warm-up steps & 0 & 0 & 500 & 500 & 0 \\
 \cmidrule(l){2-7} 
 & Optimizer & \multicolumn{5}{c}{AdamW ($\beta_1 = 0.9, \beta_2 = 0.999$)} \\
 & Gradient clip & \multicolumn{5}{c}{1} \\
 & Weight decay & \multicolumn{5}{c}{0.01} \\
\bottomrule
\end{tabular}%
}
\end{table}

\subsection{Continued Pre-training (CT)}
\label{sec:sft}

\textbf{Unified Task Sampling and Resolution Adaptability} To enhance the model's versatility across different generation paradigms, the Continued Pre-training (CT) phase adopts a unified sampling strategy that balances inputs across Text-to-Image, single-image, and multi-image tasks. This joint training regime leverages the computational budget to solidify the model's ability to handle diverse input modalities simultaneously. Furthermore, we significantly broaden the scope of our resolution management by implementing a more extensive bucket sampling protocol. We explicitly train on a comprehensive spectrum of aspect ratios—including 2:1, 16:9, 3:2, 4:3, 1:1, 3:4, 2:3, 9:16, and 1:2. This wide-ranging aspect ratio coverage prevents overfitting to standard formats and ensures the model maintains compositional integrity and visual fidelity regardless of the canvas dimensions requested by the user.

\textbf{Synthetic Augmentation and Dense Semantic Alignment} The composition of the training data during this stage is meticulously engineered to bridge the gap between quantity and quality. We maintain a curated stream of high-quality data sampled from the Internet, supplemented by a significant proportion of synthetic data designed to cover underrepresented domains. A critical innovation in this phase is the integration of dense caption training. By utilizing comprehensive, highly detailed descriptions for training samples, we compel the model to align visual features with complex and specific textual cues. This approach is particularly effective for learning long-tail vocabulary, ensuring that the model develops a robust understanding of rare objects, nuanced textures, and intricate scenarios that are often glossed over by sparse captions.

\textbf{Cluster-Based Distribution Balancing} To address the inherent imbalances found in large-scale datasets, we enforce a rigorous class-balancing protocol throughout the CT stage. Rather than relying on raw data frequency, we categorize the entire dataset into distinct semantic clusters and apply a uniform sampling strategy across these groups. This method ensures that every category—from common objects to niche artistic styles—receives sufficient exposure during training. By guaranteeing that each semantic cluster is fully and equally trained, we mitigate the risk of mode collapse or overfitting to dominant classes, resulting in a model that exhibits consistent high-quality generation capabilities across a universally diverse range of subjects.

\subsection{Supervised Fine-Tuning (SFT)}
\label{sec:sft}
\textbf{High-Fidelity Data Alignment and Balanced Distribution.} To transition the model from a broad, noisy pre-training distribution to a focused, high-quality sub-manifold, we construct a hierarchically organized dataset dominated by high-resolution imagery (1024×1024) that has undergone meticulous human filtering. Unlike the loose supervision in pre-training, our SFT phase utilizes instruction-following captions and structured prompts to enforce precise alignment between textual descriptions and visual outputs. This rigorous supervision acts as an anchor, forcing the model to discard low-quality modes (e.g., artifacts or inconsistent rendering) and converge towards superior textural quality. Furthermore, to address the risk of catastrophic forgetting regarding long-tail concepts, we enforce strict consistency across different semantic clusters. By employing a dynamic resampling strategy that ensures strict balance among data categories, we preserve the semantic diversity of the pre-trained model while elevating its aesthetic fidelity.

\textbf{Refined Optimization with Model Weight Averaging.} During the optimization process, we adopt a significantly smaller learning rate compared to the pre-training phase. This conservative update strategy is designed to refine high-frequency details and optimize photorealistic attributes without disrupting the global structural understanding established earlier. To further enhance the model's generation stability and aesthetic robustness, we incorporate an Exponential Moving Average (EMA) strategy, also referred to as Model Weight Averaging. By maintaining a moving average of model parameters throughout the training trajectory, we effectively smooth out the optimization landscape. This fusion process helps synthesize capabilities from different training steps, mitigating specific biases inherent in individual checkpoints and ensuring the final model achieves a balanced performance with greater generalization capabilities.

\subsection{Reinforcement Learning with Human Feedback (RLHF)}
\label{sec:rlhf}

\subsubsection{Direct Preference Optimization (DPO)}

We optimize the model using a DPO framework enhanced with Positive Sample Reinforcement (PSR) to ensure training stability. The alignment relies on a two-stage data synthesis strategy, combining Mix-Policy construction for robust instruction following with specific Visual-Enhancement sampling for aesthetic fidelity.

\paragraph{Positive Sample Reinforcement (PSR) Strategy} While standard DPO provides a robust framework, our experiments reveal a "double degradation" phenomenon. Specifically, we observe that the \textit{Win Diff} and \textit{Lose Diff} terms often exhibit a synchronous upward trend during training. This indicates that the optimization trajectory deviates from the high-quality data manifold; the model effectively degrades its generation capability for chosen samples ($x_w$) while attempting to distance itself from rejected ones ($x_l$).

This instability suggests that in continuous high-dimensional spaces, simply penalizing negative samples is insufficient to guide the policy toward the desired distribution. To address this, we propose an \textit{Asymmetric Gradient Optimization} strategy that anchors alignment on \textit{Positive Sample Reinforcement} (PSR). We introduce a weighting coefficient $\omega$ and an SFT regularization term $\lambda$ to explicitly prioritize the reconstruction of high-quality distributions:

\begin{equation}
\mathcal{L}_{\text{Ours}} = -\mathbb{E}_{(c,x_w, x_l) \sim \mathcal{D}} \Big[ \log \sigma \big( \beta [ \underbrace{ (\mathcal{L}_{l}^\theta - \mathcal{L}_{l}^{\text{ref}}) }_{\text{Lose Diff}} - \omega \cdot \underbrace{ (\mathcal{L}_{w}^\theta - \mathcal{L}_{w}^{\text{ref}}) }_{\text{Win Diff}} ] \big) - \lambda \mathcal{L}_{w}^\theta \Big] 
\end{equation}

By setting $\omega > 1$, we amplify the gradient contribution from the \textit{Win Diff} term, ensuring that the optimization is primarily driven by the high-fidelity modes ($x_w$) rather than being misguided by the unconstrained avoidance of negative samples.

\paragraph{Automated Data Synthesis and Mix-Policy Strategy} To bypass the costs of human annotation, we implement a fully automated pipeline that employs a \textit{Multi-Stage Instruction Evolution} strategy, using VLMs to synthesize instructions ranging from atomic tasks to complex real-world requests. To avoid the "capability ceiling" inherent in self-sampling, we adopt a \textit{Mix-Policy} data construction method. Instead of relying solely on the SFT model, we source high-quality positive samples () from diverse expert branches and aesthetic-optimized models, contrasting them with VLM-filtered negative samples () from the base model. This approach creates a steeper learning signal, breaking the self-reinforcing loop of self-generation while specifically enhancing visual fidelity and texture quality.

\subsubsection{Diffusion NFT}

In this stage, we conduct online reinforcement learning via DiffusionNFT~\cite{zheng2025diffusionnft}, optimizing a composite reward derived from the Fine-grained VLM and Layout-Aware OCR models. This process is further enhanced by a semi-hard sample mining strategy to ensure data efficiency.

\paragraph{Diffusion Negative-aware FineTuning} 
We employ DiffusionNFT to perform online reinforcement learning directly on the forward process. Distinct from DPO which requires paired data, DiffusionNFT utilizes the optimality probability $r \in [0, 1]$ of each online sampled image. The objective minimizes the weighted flow matching error via implicitly parameterized positive and negative policies:

\begin{equation}
\mathcal{L}_{\text{NFT}} = \mathbb{E}_{t, x_0 \sim \pi^{\text{old}}} \Big[ r \underbrace{ \| v_{\theta}^+(x_t, t) - v \|^2 }_{\text{Positive Match}} + (1-r) \underbrace{ \| v_{\theta}^-(x_t, t) - v \|^2 }_{\text{Negative Match}} \Big]
\end{equation}

where $v$ denotes the target velocity, and the implicit policies are defined as $v_{\theta}^+ = (1-\beta)v^{\text{old}} + \beta v_{\theta}$ and $v_{\theta}^- = (1+\beta)v^{\text{old}} - \beta v_{\theta}$. Theoretically, this formulation defines a contrastive improvement direction that pushes the policy towards high-reward regions while explicitly penalizing low-reward trajectories.

\paragraph{Fine-grained Logit-Weighted Ensembling Reward} 
Leveraging the specialized reward models detailed in Sec.~\ref{subsubsec:data-cleaning}, we derive dense reinforcement signals across critical dimensions. These models incorporate Chain-of-Thought (CoT) reasoning prior to scoring, ensuring alignment with nuanced human judgment. To circumvent the optimization instability inherent in sparse and discrete integer rewards, we adopt a continuous soft-scoring mechanism during the VLM forward pass. Specifically, immediately following the generated CoT sequence, we extract the logits corresponding to numeric rating tokens to compute a Softmax-normalized probability-weighted expectation rather than relying on hard argmax predictions. By averaging these soft scores across an ensemble of $K$ stochastic inference passes, we effectively smooth the reward landscape, guaranteeing a continuous and stable reinforcement signal $R$ for robust optimization:

\begin{equation}
    R(\mathbf{x}, \mathbf{y}) = \frac{1}{K} \sum_{k=1}^{K} \sum_{v \in \mathcal{V}} v \cdot P(v \mid \mathbf{x}, \mathbf{y}, \mathbf{c}_k), \quad \text{with} \ P(v \mid \cdot) = \frac{\exp(z_v)}{\sum_{v' \in \mathcal{V}} \exp(z_{v'})},
\end{equation}

where $\mathbf{c}_k$ denotes the CoT rationale generated in the $k$-th pass, $\mathcal{V}$ represents the set of numeric tokens (e.g., $\{1, \dots, 5\}$), and $v$ is the scalar value associated with the token.

\paragraph{Layout-Aware OCR-based Reward}
For text-editing tasks, we introduce a layout-aware OCR-based reward that more accurately reflects both textual fidelity and spatial consistency. Conventional OCR-based rewards depend solely on the edit distance between the recognized text and the target string. Although straightforward, this ignores typography and layout, allowing models to hack the reward by generating oversized characters that OCR systems recognize more easily but blend poorly with the image and disrupt layout consistency. To overcome this limitation, we incorporates layout consistency alongside text correctness. Rather than treating OCR outputs as a flat string, we decompose both predicted and reference OCR outputs into character-level elements with positions and scales, and assess whether each character appears in a plausible location and size. This formulation penalizes missing, misaligned, or abnormally scaled characters while retaining the semantic alignment signal. A lightweight gating mechanism further ensures that layout terms activate only when the text content is reasonably correct. The total reward is formulated as:
\begin{equation}
\mathcal{R}_{\text{LA-OCR}}
=
\underbrace{
w_{\text{text}}
\left(
1 -
\frac{d(s_{\text{pred}},\, s_{\text{tgt}})}{\max(|s_{\text{tgt}}|,1)}
\right)
}_{\mathcal{R}_{\text{text}}}
+
\underbrace{
w_{\text{layout}}\,
Gate(\mathcal{R}_{\text{text}})
\left(
\frac{|s_{\text{pred}}|}{|s_{\text{tgt}}|}
\sum_{i}
e^{-d_{i}}
e^{-\Delta s_{i}}
\right)
}_{\mathcal{R}_{\text{layout}}}
\end{equation}
where $s_{\text{pred}}$ and $s_{\text{tgt}}$ are the OCR-predicted and target text strings, and $d_i$ and $\Delta s_i$ denote the center-distance and overscaling penalty for the $i$-th matched character.
This design mitigates reward hacking and greatly reduces text collapse, resulting in more stable glyphs and more natural typography in text-editing tasks.

\paragraph{Semi-Hard Sample Mining for Data Efficiency}
We empirically observe that naive random sampling yields negligible gains, as the model inefficiently allocates capacity to samples already within its comfort zone. To mitigate this inefficiency, we introduce a hardness-aware curriculum that utilizes the preceding DPO checkpoint for offline hardness estimation. By generating a diverse set of candidate edits for each instruction, we estimate the underlying reward distributions. We specifically target ``semi-hard'' instances, defined by a distinct performance profile: \textit{competent yet high-variance}. These samples typically exhibit a satisfactory mean reward—indicating that the model grasps the semantic alignment—yet suffer from significant instability in their lower-bound scores. Unlike trivial samples that offer limited information gain, or completely hard samples that often precipitate optimization collapse, these semi-hard instances maximize gradient utility by targeting the model's capability boundary. By focusing on samples that are conceptually feasible but lack robustness, this strategy maximizes the information gain per training step, thereby significantly enhancing both the consistency and fidelity of the editing process.

\subsection{Consistency Loss}
We observe that maintaining identity in general-purpose image editing remains a significant challenge when relying solely on standard mean squared reconstruction losses, particularly in human-centric applications. While these objectives are effective for ensuring low-level pixel fidelity, they often fail to capture the high-level semantic nuances required for identity consistency, leading to the undesirable phenomenon of identity drift. Given the high practical value and growing demand for human-centric image editing, there is an urgent need for mechanisms that can robustly preserve identity throughout the editing process. Therefore, we integrate a robust identity consistency constraint into our training framework.

The denoising process in rectified flow follows a distinct coarse-to-fine progression. We observe that the early stage (high-noise regime) is pivotal for identity formulation, where global semantic structures are established, whereas the later stage focuses on the refinement of fine-grained textures. We argue that as the model enters the low-noise regime, the noisy input image already carries a stable identity structure. Since the identity is effectively ``locked'' within the input at this point, further identity-driven optimization is not only less likely to alter established facial features but may also introduce unintended visual artifacts by competing with the model's ability to synthesize fine-grained details. Driven by this insight, we propose a dynamic weight scheduling for the identity loss $\mathcal{L}_{id}$. Specifically, the weight $\lambda_{id}$ is defined as a function of the noise level $\sigma$:
\begin{equation}
    \lambda_{id}(\sigma) = 
    \begin{cases} 
    \eta \cdot \sigma^2, & \text{if } \sigma < 0.9 \\
    0, & \text{otherwise}
    \end{cases}
    \label{eq:weight_scheduling}
\end{equation}
where $\eta$ is a scaling hyperparameter. This quadratic decay ensures the constraint gradually tapers off as the model transitions from semantic anchoring to pixel-level refinement. The total training objective is then formulated as:
\begin{equation}
    \mathcal{L}_{total} = \mathcal{L}_{mse} + \lambda_{id}(\sigma) \cdot \mathcal{L}_{id}.
\end{equation}

The implementation of the identity loss $\mathcal{L}_{id}$ begins by defining a differentiable spatial transformation $\mathcal{T}$ designed to extract and normalize the Region of Interest (ROI) from the image. 
For human-centric editing, this transformation is specifically derived from facial landmarks detected in the ground-truth image $x_{gt}$, which aligns the face into a canonical representation. 
During training, we obtain the one-step denoised estimate $\hat{x}_0$ from the predicted velocity $v_t$ as $\hat{x}_0 = x_t - \sigma_t \cdot v_t$. 
Although $\hat{x}_0$ may be blurry or contain artifacts at higher noise levels, pre-trained semantic backbones---most notably face recognition models---exhibit significant robustness to such degradations. 
This resilience ensures that $\mathcal{L}_{id}$ can provide relative stable semantic guidance and effectively optimize the subject's identity even when the underlying image reconstruction is coarse. 
To maintain strict spatial correspondence and ensure training stability, we apply the identical transformation $\mathcal{T}$ (parameterized by $x_{gt}$) to the predicted $\hat{x}_0$, ensuring the loss is computed on perfectly aligned regions.

This framework naturally extends to multi-subject scenarios. For an image containing $N$ individuals, we detect and align each face independently to obtain a set of transformations $\{\mathcal{T}_i\}_{i=1}^N$. The total identity loss is formulated as the average cosine distance across all detected subjects:
\begin{equation}
    \mathcal{L}_{id} = \frac{1}{N} \sum_{i=1}^{N} \left( 1 - \frac{\phi(\mathcal{T}_i(\hat{x}_0)) \cdot \phi(\mathcal{T}_i(x_{gt}))}{\|\phi(\mathcal{T}_i(\hat{x}_0))\|_2 \cdot \|\phi(\mathcal{T}_i(x_{gt}))\|_2} \right),
    \label{eq:id_loss}
\end{equation}
where $\phi(\cdot)$ denotes the pre-trained face recognition backbone. By aggregating the loss across all subjects, our model effectively anchors the unique identity of each individual during the critical structural formation window while preserving textual integrity. The alignment is extensible and could potentially be adapted to the preservation of other universal objects by employing appropriate semantic encoders.


\subsection{Training Strategy.}

To enhance training stability and maximize data efficiency, our framework incorporates a rigorous optimization strategy designed to ensure robust model generalization. This approach coordinates global timestep distribution through a distributed stratified sampling mechanism and refines parameter convergence via a post-training weight averaging protocol.

\textbf{Distributed Stratified Timestep Sampling.} 
In standard distributed training, independent random sampling of diffusion timesteps across GPUs often leads to statistical clusters, preventing the global batch from uniformly covering the noise spectrum. To address this, we implement a \textit{Discrete Stratified Sampling} strategy. 
The total diffusion timestep horizon $T$ is decomposed into $K$ disjoint equidistant sub-intervals, where $K$ corresponds to the distributed world size (or group size). At each training iteration, every computational rank is assigned a specific sub-interval from which to sample $t$ uniformly. To prevent inductive bias—where specific ranks overfit to fixed noise levels—we employ a synchronized interval rotation mechanism. A global permutation seed is broadcast periodically to dynamically reassign noise intervals among ranks. This ensures that the aggregated global batch approximates a perfect uniform distribution over $t \in [0, T]$ at every step, significantly stabilizing the convergence of the diffuion objective.

\textbf{Logit-Normal Loss Weighting.} 
While the sampling probability is globally uniform, the contribution of each timestep to the gradient is modulated by a Logit-Normal weighting scheme. The loss weight is determined based on the noise level, following a distribution centered at the meaningful middle trajectory of the diffusion process. This allows the optimizer to suppress contributions from negligible high-noise or low-noise extremes while concentrating capacity on the critical intermediate phase where semantic structure and texture are primarily established.

\textbf{Model Weight Averaging.} 
To synthesize the strengths of the model across the optimization trajectory, we adopt a Exponential Moving Average (EMA) protocol. Rather than relying on a single final checkpoint, which may be sensitive to the stochasticity of the last few batches, we compute the arithmetic mean of model parameters $\theta$ over the converging iterations. This ensemble-in-weights approach smoothens the loss landscape, effectively neutralizing transient fluctuations and enhancing the model's robustness against distribution shifts in real-world editing scenarios.

\section{REDEdit-Bench}
\label{sec:rededit-benchMark}
\begin{wrapfigure}{r}{0.4\textwidth} 
    \vspace{-25pt}
    \centering
    \includegraphics[width=\linewidth]{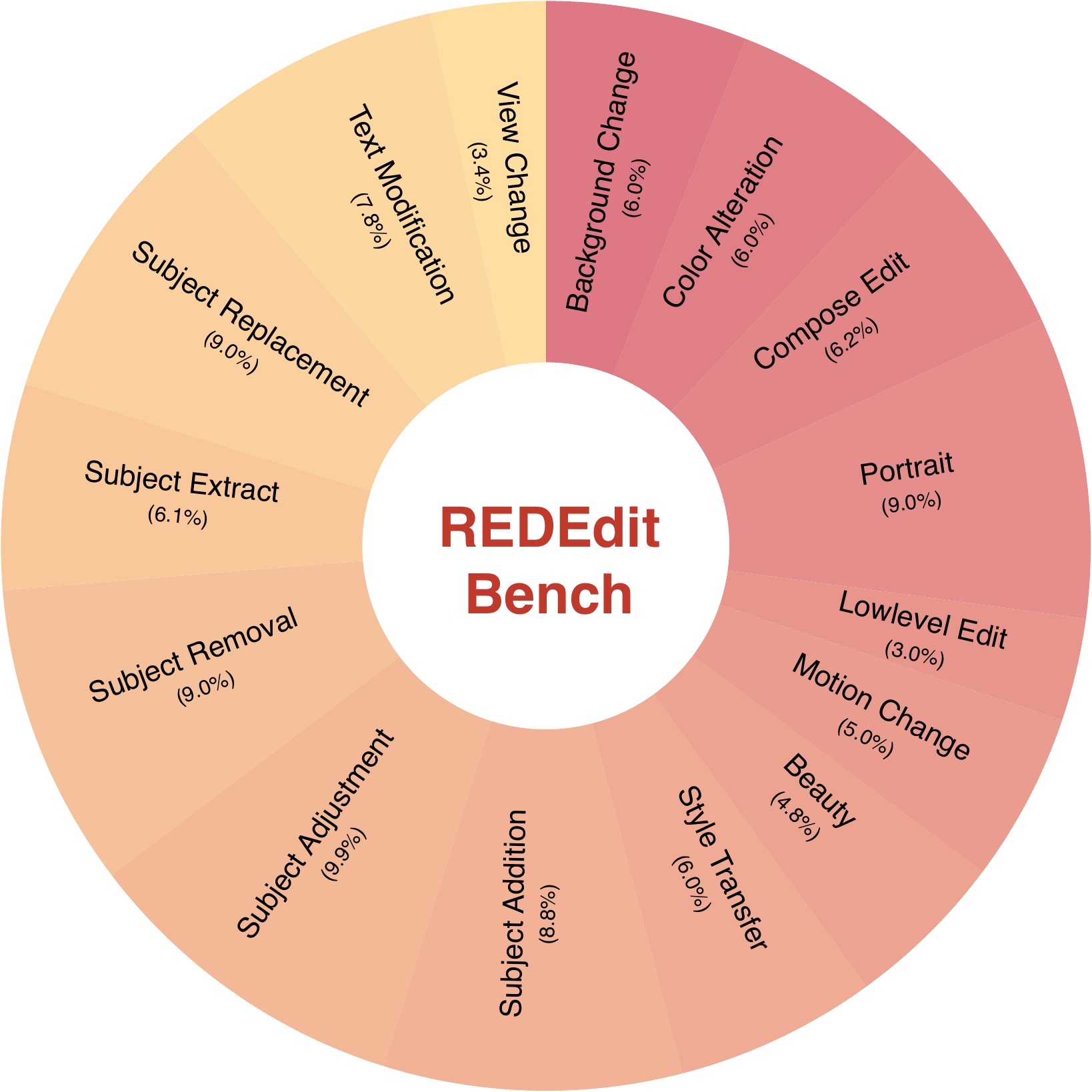}
    \caption{Task category distribution of REDEdit-Bench.}
    \label{fig:REDEdit-Bench}
    \vspace{-35pt}
\end{wrapfigure}

To better evaluate instruction-following fidelity and preservation quality, we introduce \textbf{REDEdit-Bench}, a benchmark containing 1,673 bilingual (Chinese–English) edit pairs spanning 15 structured editing categories. 
As shown in Fig.~\ref{fig:REDEdit-Bench}, the benchmark provides a balanced distribution across all task categories.
As summarized in Table~\ref{tab:benchmarks}, REDEdit-Bench demonstrates clear advantages over widely used open-source edit benchmarks in both scale and evaluation design.

\subsection{Benchmark Construction}
We collect over 3,000 real-world images from the internet and legitimate procurement channels, covering natural landscapes, architecture, objects, animals, and portraits. 
Professional technical staff first select each image and write corresponding editing instructions, ultimately constructing 1,673 bilingual (Chinese–English) edit pairs. 
These pairs are further reviewed and verified by multiple experts to ensure the diversity of the editing instructions and the legality of the image sources.

\begin{table}[h]
    \centering
    \caption{Comparison of key attributes of open-source image editing benchmarks. 
REDEdit-Bench achieves the largest scale and is the only benchmark covering all listed attributes.}

    \label{tab:benchmarks}    
    \resizebox{\textwidth}{!}{
        \begin{tabular}{lcccccc}
            \toprule
            \textbf{Benchmarks} & \textbf{Size} & \textbf{Real Image} & \textbf{Human Filtering} & \textbf{Sub-tasks} & \textbf{Bilingual} & Task-Specific Evaluation Prompts \\
            \midrule         
            MagicBrush~\cite{zhang2023magicbrush}     & 1,053 & \textcolor{green!60!black}{\ding{52}} & \textcolor{green!60!black}{\ding{52}} & 7  & \textcolor{red}{\ding{56}} & \textcolor{red}{\ding{56}} \\          AnyEdit~\cite{shen2024manytomanyimagegenerationautoregressive}        & 1,250 & 
            \textcolor{green!60!black}{\ding{52}} & 
            \textcolor{red}{\ding{56}} & 25 & 
            \textcolor{red}{\ding{56}} & 
            \textcolor{red}{\ding{56}} \\
            ImgEdit~\cite{ye2025imgedit}      & 811 & \textcolor{green!60!black}{\ding{52}} & \textcolor{green!60!black}{\ding{52}} & 14 & \textcolor{red}{\ding{56}} & \textcolor{green!60!black}{\ding{52}} \\
            GEdit-Bench~\cite{liu2025step1x-edit}    & 606   & \textcolor{green!60!black}{\ding{52}} & \textcolor{green!60!black}{\ding{52}} & 11 & \textcolor{green!60!black}{\ding{52}} & \textcolor{red}{\ding{56}} \\
            \midrule
            \textbf{REDEdit-Bench (Ours)} & 1,673 & \textcolor{green!60!black}{\ding{52}} & \textcolor{green!60!black}{\ding{52}} & 15 & \textcolor{green!60!black}{\ding{52}} & \textcolor{green!60!black}{\ding{52}} \\ 
            \bottomrule
        \end{tabular}
    }
\end{table}

\subsection{Evaluation Pipeline}
Based on REDEdit-Bench, we conduct a comprehensive evaluation of mainstream image editing models, with a primary focus on Prompt Compliance, Visual Naturalness, and Physical \& Detail Coherence. With Gemini 3 Flash employed as the automated evaluator, our model achieves SOTA performance among open-source models.

For text editing tasks, we introduce two novel metrics: OCR and VLM Judge. The OCR metric quantifies character generation precision by comparing OCR results against the ground truth to calculate the Levenshtein distance, completion rate, and word accuracy, producing a weighted normalized score. To address OCR's limitation of focusing solely on spelling, the VLM Judge utilizes Multimodal Large Language Models to evaluate visual quality. It assesses metrics including SuccessEdit, OverEdit, Style, and Consistency (background fusion). Together, these metrics construct a comprehensive evaluation balancing character accuracy with visual fidelity.


\section{Performance and Evaluation}
\label{sec:benchmark-evaluation}

We evaluate our method through both human assessment and standardized benchmark testing to provide a comprehensive analysis of its effectiveness. 
As detailed in Section~\ref{humaneval}, we first conduct a structured human evaluation where annotators assess model outputs across carefully defined dimensions. Section~\ref{qqr} then presents quantitative and qualitative results on benchmark datasets, covering four representative editing categories: general editing, text-centric editing, creative editing, and virtual try-on editing. This dual evaluation framework enables a thorough validation from both perceptual and objective perspectives.

\subsection{Human Evaluation}
\label{humaneval}
\begin{figure}[H]
    \centering
    \includegraphics[width=0.85\textwidth]{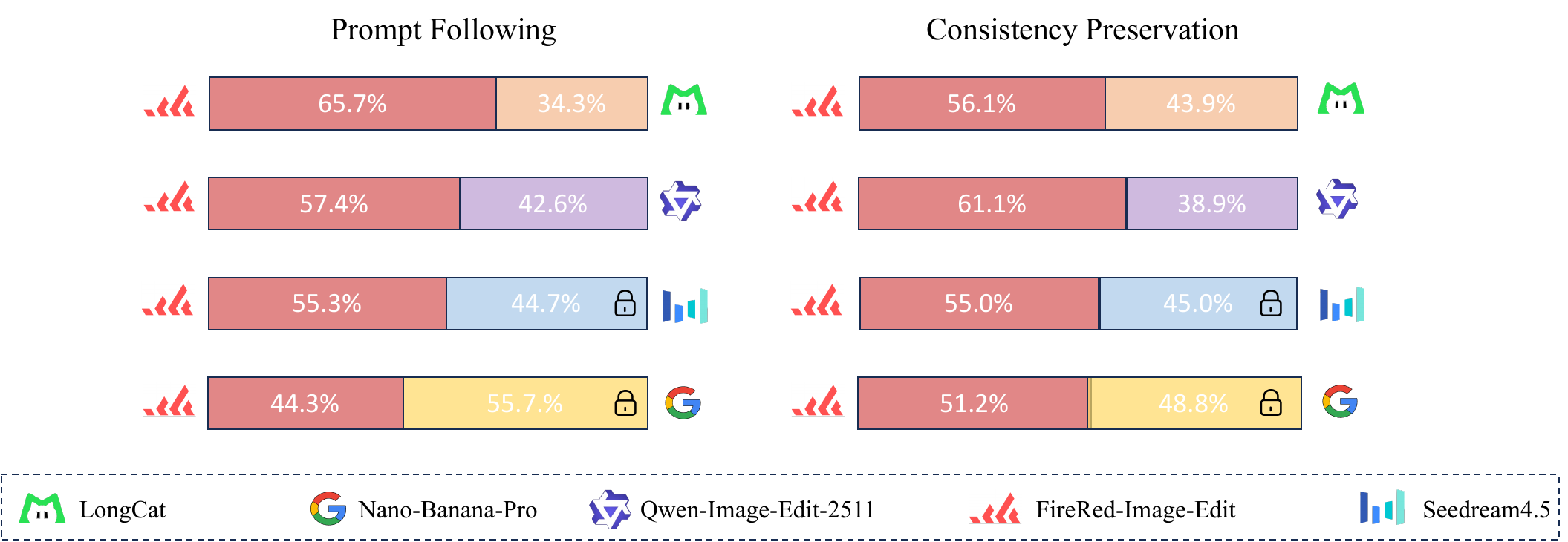}
    \caption{The figure compares FireRed-Image-Edit against four other models across two metrics. FireRed-Image-Edit achieves the highest consistency scores and leads in text-to-image alignment against most competitors, except for Nano-Pro which slightly outperforms it in that category.}
    \label{fig:human_eval}
\end{figure}

To comprehensively assess the perceptual quality and practical effectiveness of our image editing foundation model, we conduct a multi-model blind human evaluation against strong open-source and commercial baselines. Instead of pairwise comparison, multiple model outputs for the same input are presented simultaneously in randomized order without revealing model identities. Annotators independently evaluate each result to ensure fairness and eliminate positional or brand bias. This evaluation protocol is designed to jointly measure the correctness of instruction execution and the perceptual quality of the edited results from multiple complementary perspectives.

\subsubsection{Evaluation Dimensions}
We evaluate each editing result along two carefully designed dimensions:

\textbf{Prompt Following.} This dimension measures how accurately and completely the model understands and executes the user-provided prompt. Annotators assess whether the intended edit type is correctly identified, all key constraints are satisfied, and the final output faithfully reflects the user’s editing intent without missing or extraneous modifications.
    
\textbf{Consistency Preservation.} This metric focuses on the protection of non-edited regions. Ideally, only regions explicitly specified by the prompt should be modified, while all other content—such as facial identity, background, object structure, texture, and global layout—remains unchanged. This dimension is particularly critical for localized and iterative editing scenarios.
    
    

\subsubsection{Results Analysis}
As illustrated in Fig.~\ref{fig:human_eval}, our model achieves leading performance across \textbf{Prompt Following} and \textbf{Consistency}. \methodname significantly outperforms open-source baselines such as LongCat and Qwen-Image-Edit-2511, and remains competitive with commercial systems, indicating strong instruction understanding and execution capability. 

For \textbf{Consistency Preservation}, our model obtains the highest score among all compared methods, demonstrating superior ability to modify only the intended regions while preserving non-edited content. This highlights its robustness in controlled and precise editing scenarios.

\begin{table}[t]
\centering
\caption{Results on ImgEdit-Bench (Category-wise and Overall Performance).
Best results are shown in bold and second-best results are underlined.}
\label{tab:imgedit_benchmark}
\begingroup
\setlength{\tabcolsep}{2.2pt}
\renewcommand{\arraystretch}{1.05}
\scriptsize
\begin{adjustbox}{width=\textwidth,center}
\begin{tabular}{l|ccccccccc|c}
\toprule
\textbf{Model} & \textbf{Add} & \textbf{Adjust} & \textbf{Extract} & \textbf{Replace} & \textbf{Remove} & \textbf{Background} & \textbf{Style} & \textbf{Hybrid} & \textbf{Action} & \textbf{Overall}$\uparrow$ \\
\midrule
Nano-Banana~\cite{nanopro}
& \textbf{4.62} & 4.41 & \underline{3.68} & 4.34 & 4.39 & \textbf{4.40} & 4.18 & \textbf{3.72} & \textbf{4.83} & 4.29 \\

Seedream4.0~\cite{seedream2025seedream}
& 4.33 & 4.38 & \textbf{3.89} & \underline{4.65} & \underline{4.57} & 4.35 & 4.22 & \underline{3.71} & 4.61 & 4.30 \\

Seedream4.5~\cite{seedream2025seedream}
& \underline{4.57} & \textbf{4.65} & 2.97 & \textbf{4.66} & 4.46 & \underline{4.37} & \underline{4.92} & \underline{3.71} & 4.56 & \underline{4.32} \\

Nano-Banana-Pro~\cite{nanopro}
& 4.44 & \underline{4.62} & 3.42 & 4.60 & \textbf{4.63} & 4.32 & \textbf{4.97} & 3.64 & \underline{4.69} & \textbf{4.37} \\

\midrule
Instruct-Pix2Pix~\cite{brooks2023instructpix2pix}  
& 2.45 & 1.83 & 1.44 & 2.01 & 1.50 & 1.44 & 3.55 & 1.20 & 1.46 & 1.88 \\

MagicBrush~\cite{zhang2023magicbrush}  
& 2.84 & 1.58 & 1.51 & 1.97 & 1.58 & 1.75 & 2.38 & 1.62 & 1.22 & 1.90 \\

AnyEdit~\cite{yu2025anyedit} 
& 3.18 & 2.95 & 1.88 & 2.47 & 2.23 & 2.24 & 2.85 & 1.56 & 2.65 & 2.45 \\

UltraEdit~\cite{zhao2024ultraedit}  
& 3.44 & 2.81 & 2.13 & 2.96 & 1.45 & 2.83 & 3.76 & 1.91 & 2.98 & 2.70 \\

OmniGen~\cite{xiao2025omnigen}  
& 3.47 & 3.04 & 1.71 & 2.94 & 2.43 & 3.21 & 4.19 & 2.24 & 3.38 & 2.96 \\

ICEdit~\cite{zhang2025icedit}  
& 3.58 & 3.39 & 1.73 & 3.15 & 2.93 & 3.08 & 3.84 & 2.04 & 3.68 & 3.05 \\

BAGEL~\cite{deng2025bagel}  
& 3.56 & 3.31 & 1.70 & 3.30 & 2.62 & 3.24 & 4.49 & 2.38 & 4.17 & 3.20 \\

UniWorld-V1~\cite{lin2025uniworld}  
& 3.82 & 3.64 & 2.27 & 3.47 & 3.24 & 2.99 & 4.21 & 2.96 & 2.74 & 3.26 \\

OmniGen2~\cite{wu2025omnigen2}  
& 3.57 & 3.06 & 1.77 & 3.74 & 3.20 & 3.57 & 4.81 & 2.52 & 4.68 & 3.44 \\

Dreamomini2~\cite{xia2025dreamomni2} 
& 3.93 & 3.09 & 2.11 & 3.95 & 3.64 & 3.75 & 4.38 & 2.90 & 4.04 & 3.53 \\

FLUX.1 Kontext [Dev]~\cite{labs2025flux} 
& 3.99 & 3.88 & 2.19 & 4.27 & 3.13 & 3.98 & 4.51 & 3.23 & 4.18 & 3.71 \\

Step1X-Edit-v1.2~\cite{liu2025step1x}  
& 3.91 & 4.04 & 2.68 & 4.48 & 4.26 & 3.90 & 4.82 & 3.23 & 4.22 & 3.95 \\

Qwen-Image-Edit-2509~\cite{qwenimage}
& 4.34 & 4.27 & 3.42 & 4.73 & 4.36 & 4.37 & 4.91 & 3.56 & 4.80 & 4.31 \\

FLUX.2 [Dev]~\cite{flux-2-2025}
& 4.50 & 4.18 & 3.83 & 4.65 & \underline{4.65} & 4.31 & 4.88 & 3.46 & 4.70 & 4.35 \\

Emu3.5~\cite{cui2025emu35nativemultimodalmodels}
& \textbf{4.61} & 4.32 & 3.96 & \textbf{4.84} & 4.58 & 4.35 & 4.79 & 3.69 & 4.57 & 4.41 \\

ChronoEdit~\cite{wu2025chronoedit}
& 4.48 & 4.39 & 3.49 & 4.66 & \textbf{4.67} & \textbf{4.57} & 4.91 & 3.82 & \textbf{4.83} & 4.42 \\

LongCat-Image-Edit~\cite{LongCat-Image}
& 4.44 & 4.53 & 3.83 & \underline{4.80} & 4.60 & 4.33 & \underline{4.92} & 3.75 & \underline{4.82} & 4.45 \\

Qwen-Image-Edit-2511~\cite{qwenimage}
& 4.54 & \underline{4.57} & \underline{4.13} & 4.70 & 4.46 & 4.36 & 4.89 & \textbf{4.16} & 4.81 & \underline{4.51} \\

\rowcolor{red!10}
\textbf{\methodname} 
& \underline{4.55} & \textbf{4.66} & \textbf{4.34} & 4.75 & 4.58 & \underline{4.45} & \textbf{4.97} & \underline{4.07} & 4.71 & \textbf{4.56} \\
\bottomrule
\end{tabular}
\end{adjustbox}
\endgroup
\end{table}

\begin{table}[h!]
\centering
\caption{Results on GEdit (Category-wise and Overall Performance). Best results are shown in bold and second-best results are underlined.}
\label{tab:gedit_benchmark}
\begin{tabular}{l|ccc|ccc}
\toprule
\multirow{2}{*}{\textbf{Model}} & \multicolumn{3}{c|}{\textbf{GEdit-Bench-EN}} & \multicolumn{3}{c}{\textbf{GEdit-Bench-CN}} \\
\cmidrule(lr){2-4} \cmidrule(lr){5-7}
& \textbf{G\_SC}$\uparrow$ & \textbf{G\_PQ}$\uparrow$ & \textbf{G\_O}$\uparrow$ 
& \textbf{G\_SC}$\uparrow$ & \textbf{G\_PQ}$\uparrow$ & \textbf{G\_O}$\uparrow$ \\
\midrule

Nano-Banana~\cite{nanopro} 
& 7.396 & \textbf{8.454} & 7.291
& 7.540 & \textbf{8.424} & 7.399 \\

Seedream4.0~\cite{seedream2025seedream} 
& \underline{8.143} & 8.124 & 7.701 
& \underline{8.159} & 8.074 & 7.692 \\

Nano-Banana-Pro~\cite{nanopro} 
& 8.102 & \underline{8.344} & \underline{7.738} 
& 8.135 & \underline{8.306} & \underline{7.799} \\

Seedream4.5~\cite{seedream2025seedream} 
& \textbf{8.268} & 8.167 & \textbf{7.820}
& \textbf{8.254} & 8.167 & \textbf{7.800} \\

\midrule

FLUX.2 [Dev]~\cite{flux-2-2025} 
& 7.835 & 8.064 & 7.413 
& 7.697 & 8.046 & 7.278 \\

Qwen-Image-Edit-2509~\cite{qwenimage} 
& 7.974 & 7.714 & 7.480 
& 7.988 & 7.679 & 7.467 \\

Step1X-Edit-v1.2~\cite{liu2025step1x} 
& 7.974 & 7.714 & 7.480 
& 7.988 & 7.679 & 7.467 \\

Longcat-Image-Edit~\cite{LongCat-Image}
& 8.128 & 8.177 & 7.748 
& 8.141 & 8.117 & 7.731 \\

Qwen-Image-Edit-2511~\cite{qwenimage} 
& \underline{8.297} & \underline{8.202} & \underline{7.877} 
& \underline{8.252} & \underline{8.134} & \underline{7.819} \\

\rowcolor{red!10} \textbf{\methodname} 
& \textbf{8.363} & \textbf{8.245} & \textbf{7.943} 
& \textbf{8.287} & \textbf{8.227} & \textbf{7.887} \\

\bottomrule
\end{tabular}
\end{table}

\subsection{Quantitative and Qualitative Results}
\label{qqr}
\subsubsection{General Editing}
\paragraph{ImgEdit.}
\label{sec:imgedit}
We evaluate our model on the ImgEdit benchmark, which measures instruction completion and visual quality.
Table~\ref{tab:imgedit_benchmark} shows that \methodname achieves state-of-the-art performance, surpassing both open-source and closed-source models.
This result highlights the model's proficiency in following common editing commands.


\paragraph{GEdit.}
\label{sec:gedit}
We assess our model's capabilities on the GEdit benchmark, a comprehensive evaluation suite designed to test both instruction adherence and output visual fidelity. As illustrated in Table~\ref{tab:gedit_benchmark}, \methodname outperforms existing open-source models. This superior performance underscores the model's exceptional command comprehension and precise execution of diverse editing tasks.

\paragraph{REDEdit-Bench.}
\label{sec:redbench_results}
REDEdit-Bench is our self-built benchmark to evaluate real-world editing usage, with stronger emphasis on complex composite instructions and preservation constraints.
It is constructed in Section~\ref{sec:rededit-benchMark}. 
Table~\ref{tab:redbench_general_from_img_cn}-~\ref{tab:redbench_general_from_img_en} summarizes the main results. \methodname ranks highest among open-source models, demonstrating competitive instruction-based editing performance.

\begin{table}[t]
\centering
\caption{Results on REDEdit-Bench-CN (General dimensions).
Best results are shown in bold and second-best results are underlined.}
\label{tab:redbench_general_from_img_cn}
\begingroup
\setlength{\tabcolsep}{2.6pt}
\renewcommand{\arraystretch}{1.10}
\scriptsize
\begin{adjustbox}{width=\textwidth,center}
\begin{tabular}{l|cccccccccccccccc}
\toprule
\textbf{Model} &
\textbf{Overall} &
\textbf{Add} &
\textbf{Adjust} &
\textbf{BG} &
\textbf{Beauty} &
\textbf{Color} &
\textbf{Compose} &
\textbf{Extract} &
\textbf{Portrait} &
\textbf{Low-level} &
\textbf{Motion} &
\textbf{Remove} &
\textbf{Replace} &
\textbf{Stylize} &
\textbf{Text} &
\textbf{Viewpoint} \\
\midrule
Nano-Banana~\cite{nanopro}
& 4.13 & \textbf{4.66} & \underline{4.26} & \textbf{4.63} & \textbf{4.37} & 4.13 & 3.94 & \underline{3.17} & \underline{4.83} & 4.05 & 4.75 & 4.07 & \underline{4.74} & 3.63 & 3.69 & 3.09 \\
Seedream4.0~\cite{seedream2025seedream}
& 4.15 & 4.55 & 4.11 & \underline{4.61} & 3.83 & \underline{4.14} & \underline{4.16} & 2.48 & 4.77 & \underline{4.17} & 4.68 & 4.02 & 4.53 & \textbf{4.94} & 3.94 & 3.29 \\
Seedream4.5~\cite{seedream2025seedream}
& \underline{4.18} & \underline{4.58} & 4.09 & 4.57 & 3.97 & 4.12 & 4.05 & 2.56 & 4.80 & 3.99 & \underline{4.78} & \underline{4.12} & 4.53 & \textbf{4.94} & \underline{4.07} & \underline{3.53} \\
Nano-Banana-Pro~\cite{nanopro}
& \textbf{4.48} & \textbf{4.66} & \textbf{4.41} & 4.58 & \underline{4.35} & \textbf{4.58} & \textbf{4.36} & \textbf{3.42} & \textbf{4.86} & \textbf{4.46} & \textbf{4.91} & \textbf{4.54} & \textbf{4.79} & \underline{4.85} & \textbf{4.69} & \textbf{3.75} \\

\midrule

Qwen-Image-Edit-2509~\cite{qwenimage}
& 4.00 & 4.45 & 4.04 & 4.48 & 3.36 & \underline{4.20} & 3.92 & 2.64 & 4.16 & 3.52 & 4.66 & 4.27 & 4.66 & 4.81 & 3.53 & 3.32 \\

FLUX.2 [Dev]~\cite{flux-2-2025}
& 4.05 & 4.31 & 3.88 & \underline{4.57} & \textbf{3.80} & 3.91 & 3.85 & 2.47 & \textbf{4.50} & \underline{4.43} & \textbf{4.68} & 3.50 & 4.47 & \textbf{4.95} & 3.53 & \textbf{3.88} \\

Longcat-Image-Edit~\cite{LongCat-Image}
& 4.12 & 4.34 & \underline{4.25} & 4.54 & \underline{3.72} & 4.12 & 3.92 & 2.48 & \underline{4.49} & 4.31 & \underline{4.67} & 4.27 & 4.61 & \underline{4.94} & 3.83 & 3.30 \\

Qwen-Image-Edit-2511~\cite{qwenimage}
& \underline{4.18} & \underline{4.50} & 4.23 & 4.52 & 3.61 & 4.09 & \underline{4.00} & \underline{3.22} & 4.31 & 4.19 & 4.66 & \underline{4.41} & \underline{4.68} & 4.83 & \underline{4.08} & \underline{3.51} \\

\rowcolor{red!10}
\textbf{\methodname}
& \textbf{4.33} & \textbf{4.57} & \textbf{4.37} & \textbf{4.64} & 3.69 & \textbf{4.45} & \textbf{4.29} & \textbf{3.49} & \textbf{4.50} & \textbf{4.56} & 4.65 & \textbf{4.47} & \textbf{4.81} & 4.93 & \textbf{4.49} & 3.14 \\

\bottomrule
\end{tabular}
\end{adjustbox}
\endgroup
\end{table}

\begin{table}[t]
\centering
\caption{Results on REDEdit-Bench-EN (General dimensions). Best results are shown in bold and second-best results are underlined.}
\label{tab:redbench_general_from_img_en}
\begingroup
\setlength{\tabcolsep}{2.6pt}
\renewcommand{\arraystretch}{1.10}
\scriptsize
\begin{adjustbox}{width=\textwidth,center}
\begin{tabular}{lcccccccccccccccc}
\toprule
\textbf{Model} &
\textbf{Overall} &
\textbf{Add} &
\textbf{Adjust} &
\textbf{BG} &
\textbf{Beauty} &
\textbf{Color} &
\textbf{Compose} &
\textbf{Extract} &
\textbf{Portrait} &
\textbf{Low-level} &
\textbf{Motion} &
\textbf{Remove} &
\textbf{Replace} &
\textbf{Stylize} &
\textbf{Text} &
\textbf{Viewpoint} \\
\midrule

Nano-Banana~\cite{nanopro}
& 4.15 & 4.65 & \underline{4.23} & 4.60 & \textbf{4.37} & 4.08 & 3.98 & \textbf{3.39} & 4.72 & 4.03 & 4.63 & 4.07 & \underline{4.68} & 3.68 & 3.87 & 3.23 \\

Seedream4.0~\cite{seedream2025seedream}
& 4.18 & 4.59 & 4.12 & \underline{4.63} & 3.89 & \underline{4.10} & \underline{4.14} & 2.28 & \underline{4.77} & 4.12 & 4.73 & 4.23 & 4.56 & \textbf{4.98} & \underline{4.21} & 3.42 \\

Seedream4.5~\cite{seedream2025seedream}
& \underline{4.20} & \underline{4.66} & 4.08 & \textbf{4.64} & \underline{4.12} & 4.07 & 4.10 & 2.23 & 4.74 & \underline{4.28} & \underline{4.75} & \underline{4.24} & 4.58 & \underline{4.97} & 4.20 & \underline{3.44} \\

Nano-Banana-Pro~\cite{nanopro}
& \textbf{4.42} & \textbf{4.72} & \textbf{4.40} & \textbf{4.64} & \textbf{4.37} & \textbf{4.43} & \textbf{4.32} & \underline{3.25} & \textbf{4.82} & \textbf{4.36} & \textbf{4.85} & \textbf{4.52} & \textbf{4.75} & 4.90 & \textbf{4.54} & \textbf{3.51} \\

\midrule

Qwen-Image-Edit-2509~\cite{qwenimage}
& 3.99 & \underline{4.47} & 4.06 & 4.49 & 3.13 & 3.98 & 3.85 & 2.91 & 4.30 & 3.71 & 4.58 & 4.40 & \underline{4.67} & 4.77 & 3.77 & 2.85 \\

FLUX.2 [Dev]~\cite{flux-2-2025}
& 4.07 & 4.37 & 3.96 & 4.47 & \underline{3.72} & 3.86 & 3.87 & 2.36 & 4.44 & \underline{4.45} & 4.67 & 4.02 & 4.48 & \underline{4.87} & 3.80 & \textbf{3.84} \\

LongCat-Image-Edit~\cite{LongCat-Image}
& 4.12 & 4.38 & 4.04 & 4.49 & \textbf{3.89} & \underline{4.10} & 3.93 & 2.98 & \underline{4.47} & 4.27 & \underline{4.69} & 4.24 & 4.51 & 4.86 & 3.83 & 3.25 \\

Qwen-Image-Edit-2511~\cite{qwenimage}
& \underline{4.23} & \textbf{4.55} & \underline{4.17} & \underline{4.56} & 3.49 & 4.07 & \underline{4.07} & \textbf{3.54} & 4.42 & \textbf{4.52} & \textbf{4.72} & \underline{4.42} & 4.65 & 4.85 & \underline{4.06} & \underline{3.38} \\

\rowcolor{red!10}
\textbf{\methodname}
& \textbf{4.26} & 4.41 & \textbf{4.33} & \textbf{4.60} & 3.55 & \textbf{4.47} & \textbf{4.25} & \underline{3.49} & \textbf{4.50} & 4.44 & 4.65 & \textbf{4.46} & \textbf{4.70} & \textbf{4.94} & \textbf{4.44} & 2.78 \\

\bottomrule
\end{tabular}
\end{adjustbox}
\endgroup
\end{table}

\begin{table}[t]
\centering
\caption{Results on REDEdit-Bench (Text dimension).}
\label{tab:redbench_text}
\begingroup
\setlength{\tabcolsep}{1.8pt}
\renewcommand{\arraystretch}{0.95}
\scriptsize
\begin{adjustbox}{width=0.6\textwidth,center}
\begin{tabular}{l|ccccc}
\toprule
\textbf{Model}  & \textbf{OCR} & \textbf{SuccessEdit} & \textbf{OverEdit} & \textbf{Style} & \textbf{Consistency}  \\
\midrule
Nano-Banana~\cite{nanopro}
& 0.958 & 7.93 & \underline{9.53} & \underline{9.30} & 8.04 \\

Seedream4~\cite{seedream2025seedream}
& 0.969 & 8.59 & 9.26 & 9.19 & \underline{9.15} \\

Seedream4.5~\cite{seedream2025seedream}
& \underline{0.975} & \underline{8.61} & 9.37 & 9.00 & 8.93 \\

Nano-Banana-Pro~\cite{nanopro}
& \textbf{0.984} & \textbf{9.54} & \textbf{9.63} & \textbf{9.68} & \textbf{9.53} \\

\midrule

FLUX.2 [Dev]~\cite{flux-2-2025}
& 0.950 & 7.93 & 8.90 & 8.83 & 8.33 \\
Qwen-Image-Edit-2509~\cite{qwenimage}
& 0.957 & 8.35 & 9.26 & 9.15 & 8.15 \\
Qwen-Image-Edit-2511~\cite{qwenimage}
& 0.969 & \underline{9.42} & \underline{9.36} & \underline{9.31} & \underline{9.27} \\
LongCat-Image-Edit~\cite{LongCat-Image}
& \underline{0.976} & 8.60 & 8.96 & 8.67 & 8.50 \\
\rowcolor{red!10}
\textbf{\methodname}
& \textbf{0.983} & \textbf{9.57} & \textbf{9.53} & \textbf{9.49} & \textbf{9.51} \\

\bottomrule
\end{tabular}
\end{adjustbox}
\endgroup
\end{table}

\begin{figure}[H]
    \centering
    \includegraphics[width=0.85\textwidth]{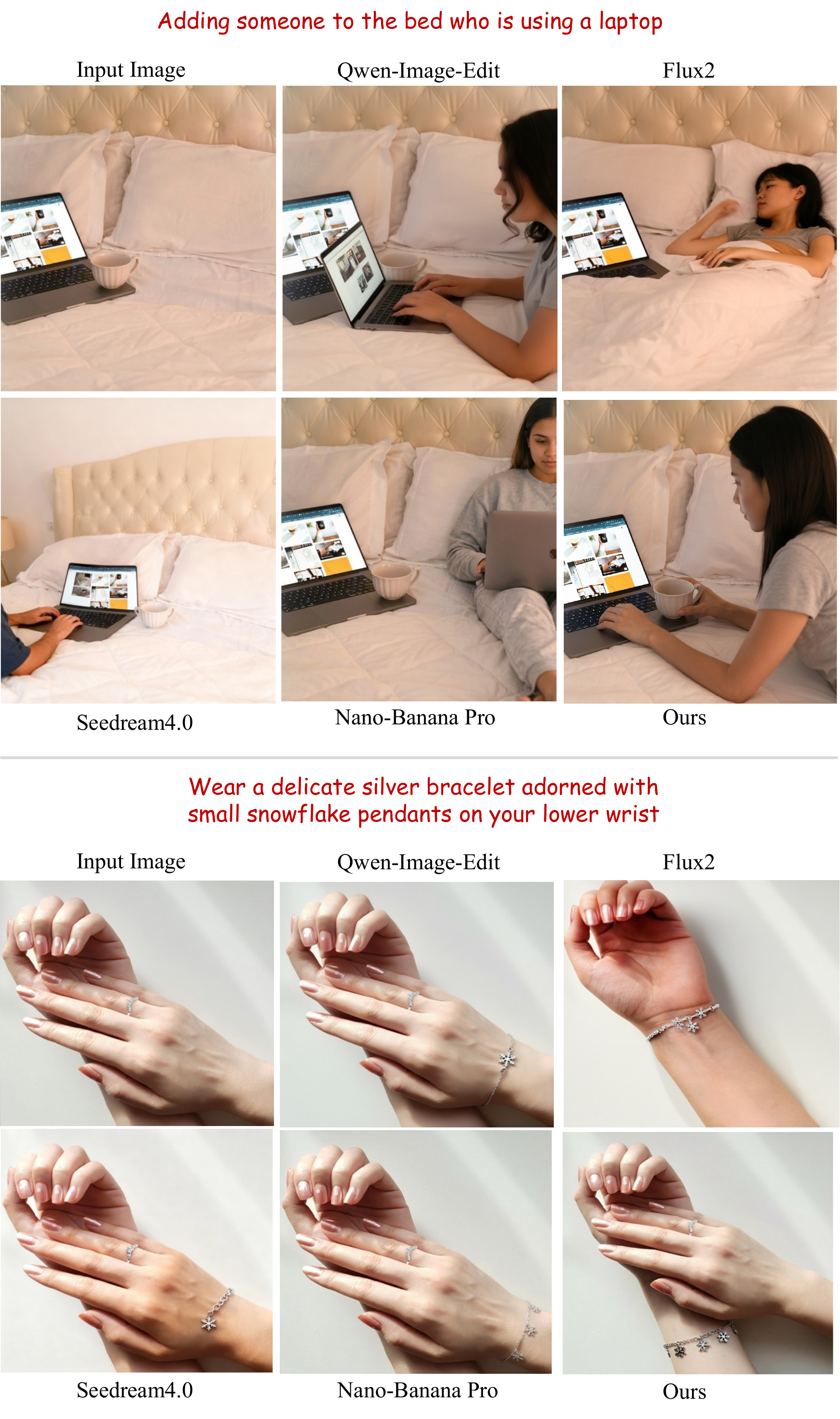}
    \caption{Demonstration of Object Addition, where the model seamlessly inserts new elements into the scene, maintaining consistency with the original context. Our model outperforms competitors by strictly adhering to instructions and ensuring the added objects blend naturally with the scene.}
    \label{fig:showcase_object_addition}
\end{figure}
\begin{figure}[H]
    \centering
    \includegraphics[width=0.8\textwidth]{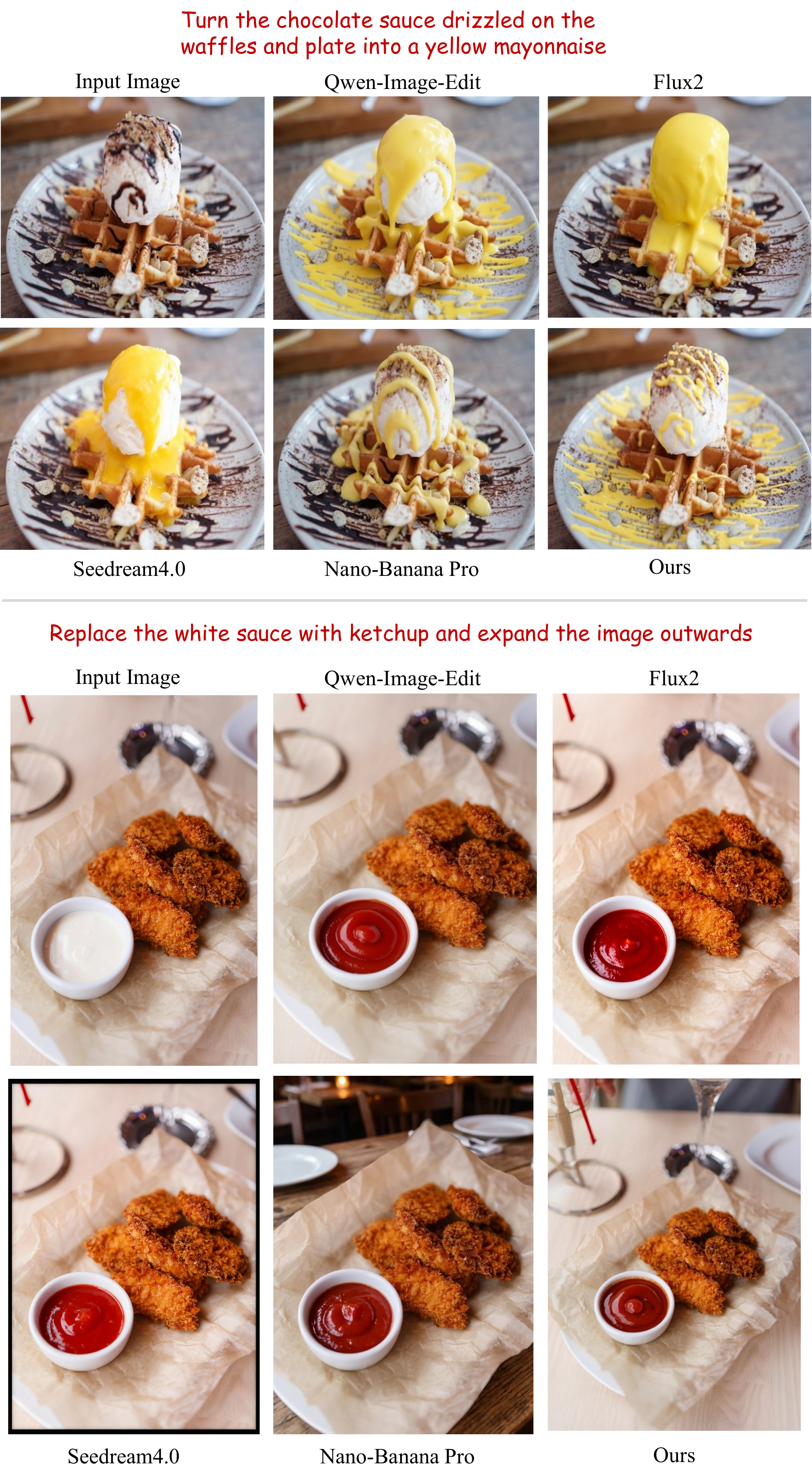}
    \caption{Examples of Object Modification, demonstrating precise control over object attributes. Our model achieves accurate editing without disrupting the surrounding textures, whereas competitors often over-edit the subject or lose detail in non-target areas.}
    \label{fig:showcase_object_modification}
\end{figure}
\begin{figure}[H]
    \centering
    \includegraphics[width=0.76\textwidth]{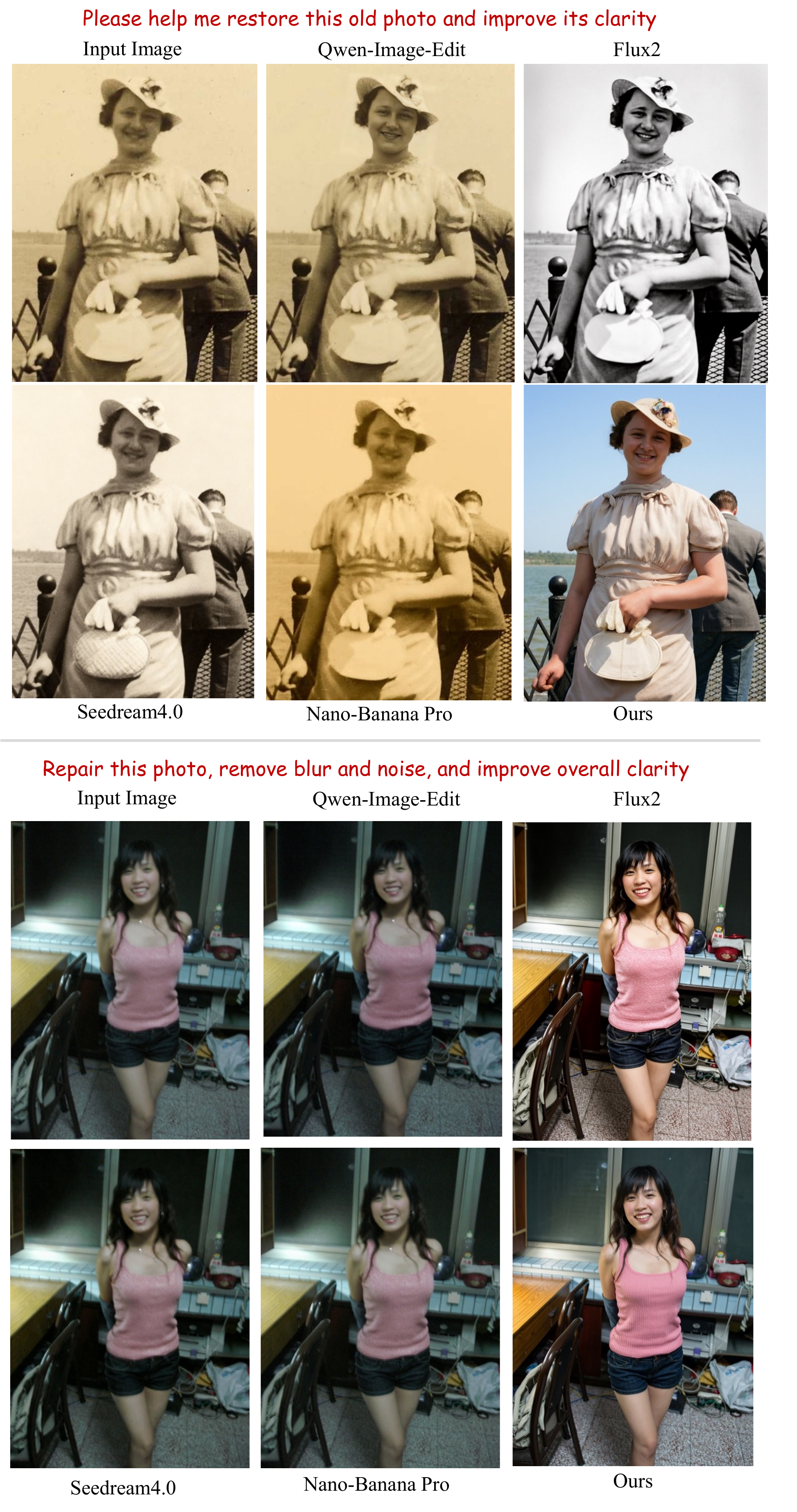}
    \caption{Cases of Low-Level Editing tasks, demonstrating the model's capability in image restoration and enhancement. Our model significantly outperforms baselines in detail recovery and visual quality, effectively reconstructing high-fidelity details from heavily degraded inputs.}
    \label{fig:showcase_low_level_edit}
\end{figure}

\paragraph{Visualization}
We show general-purpose editing cases that correspond to frequently encountered and well-defined user requests. These examples include object insertion and replacement (\emph{add}, \emph{replace}), attribute and appearance modification (\emph{color}), hybrid edit tasks (\emph{compose}), as well as image restoration tasks (denoted as \emph{low-level}).
Such tasks require precise alignment between the instruction and the results, while maintaining global consistency and avoiding unnecessary changes to unrelated regions.
The presented Figs.~\ref{fig:showcase_object_addition}--\ref{fig:showcase_low_level_edit} demonstrate that \methodname can reliably perform these edits with accurate localization, physically plausible interactions, and stable visual quality, making it suitable for a wide range of everyday image editing applications.

\subsubsection{Text-Centric Editing}

\paragraph{Text-Centric Evaluation.}
To further evaluate text editing capabilities, we apply our proposed OCR and VLM Judge pipeline to the text subset of REDEdit-Bench, assessing performance across five specific dimensions.
The evaluation examines the accuracy of edited content at the character and word level, the preservation of original font style and visual attributes, the overall visual consistency between the edited text and the surrounding image, and whether unnecessary modifications are introduced beyond the given instruction. This setting enables a more fine-grained analysis of text-centric behavior. 
As shown in Table~\ref{tab:redbench_text}, \methodname achieves leading results among open-source models, demonstrating strong capability in accurate text modification while maintaining style fidelity and global coherence.

\paragraph{Visualization.}
We additionally present text-centric editing results in Figs.~\ref{fig:text_showcase}--\ref{fig:poster_showcase} focusing on complex text generation and poster-style visual design.
The showcased examples include long-form text synthesis with consistent typography and layout, precise text replacement and correction while preserving font style and visual appearance, as well as poster and cover editing scenarios where text content must be harmoniously integrated with the underlying imagery.
These cases require accurate instruction following at the character and word level, together with global layout awareness and aesthetic consistency across text, graphics, and background.
The results demonstrate that \methodname can reliably handle both content correctness and visual coherence in text-heavy editing scenarios, making it suitable for practical applications such as poster design, cover editing, and creative visual communication.

\subsubsection{Creative Editing}
Creative editing differs from  image editing tasks in that it often requires deeper instruction understanding, implicit scene reasoning, and global compositional restructuring. The challenge lies in synthesizing content that follows the text but remains visually consistent with the original image.

\paragraph{Visualization.}
Fig.~\ref{fig:showcase_creativity} demonstrates the creative editing capabilities of \methodname. 
The displayed results range from structural abstraction (e.g., design sketches, cross-sectional views) and concept-driven synthesis (e.g., sculptural styles, foldable designs) to imaginative edits that defy physical constraints (e.g., floating objects). 
While non-exhaustive, these examples demonstrate that \methodname can effectively translate high-level creative intents into coherent visual outcomes.


\subsubsection{Try-on Editing}
We further evaluate \methodname under structured virtual try-on settings, where the model is tasked with transfer garments from a reference image onto a target person while simultaneously following detailed styling instructions.

\begin{figure}[H]
    \centering
    \includegraphics[width=1.0\textwidth]{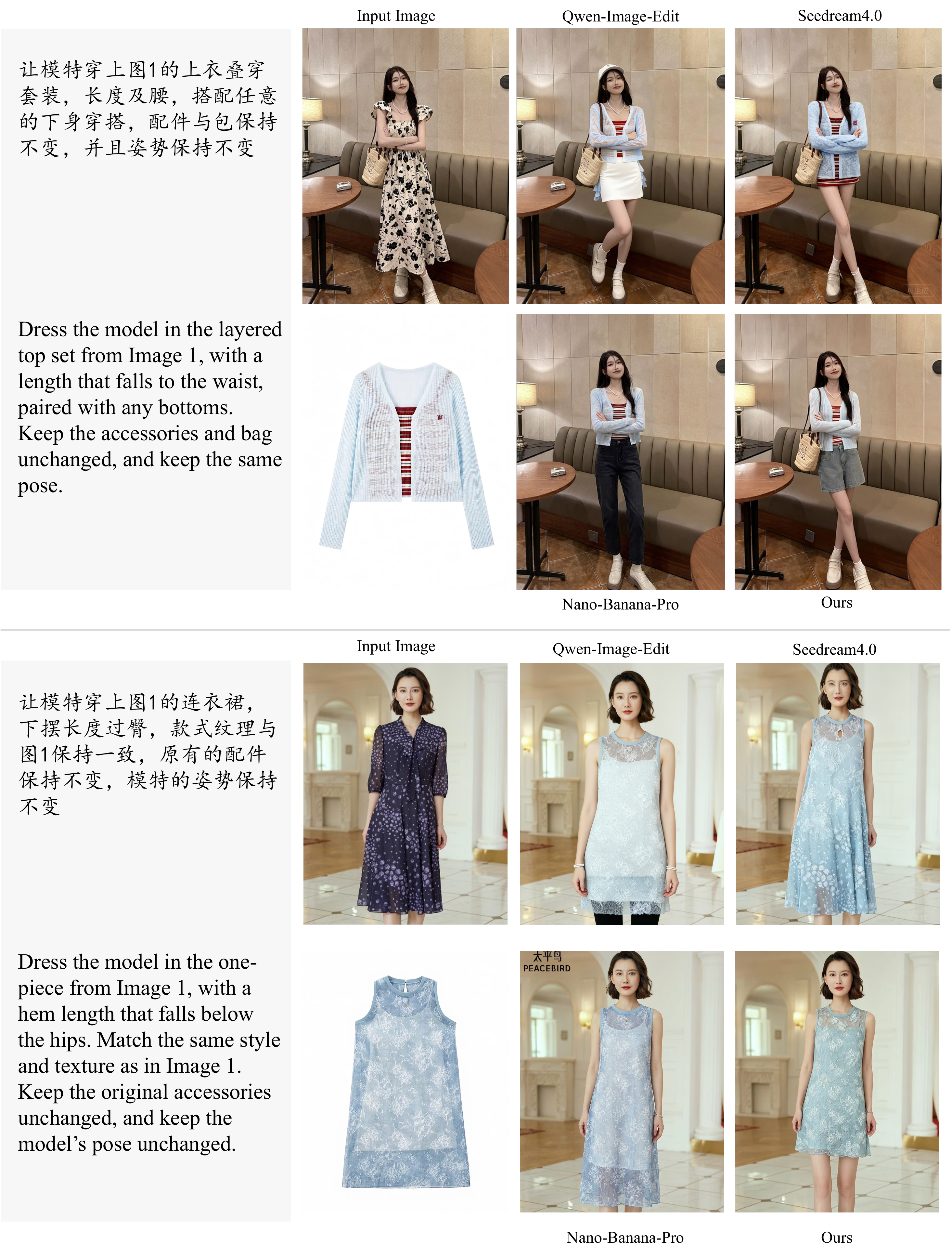}
    \caption{Visualization of the Virtual Try-on task, where our model transfers reference garments with superior fidelity compared to competitors, accurately rendering text-specified styling attributes (e.g., fit, accessories) while preserving details that baselines often distort or overlook.}
    \label{fig:showcase_virtual_tryon}
\end{figure}

\begin{figure}[H]
  \centering
  \includegraphics[width=1\textwidth]{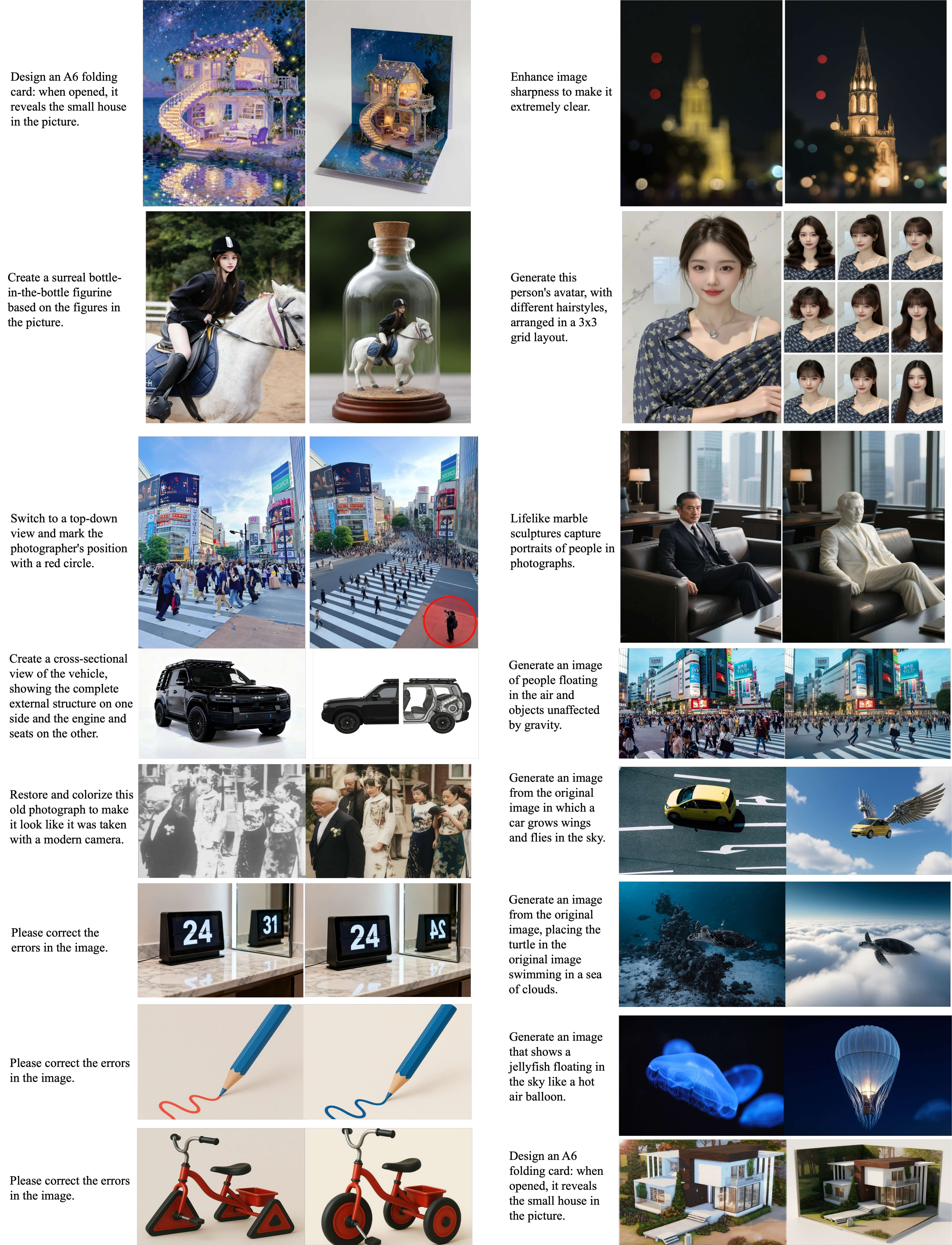}
  \caption{This visualization demonstrates the power of our FireRed-Image-Edit in a variety of creative scenarios. The results clearly demonstrate the model's ability to interpret complex instructions and generate high-quality, creative edited images.}
  \label{fig:showcase_creativity}
\end{figure}

\begin{figure}[H]
    \centering
    \includegraphics[width=\textwidth]{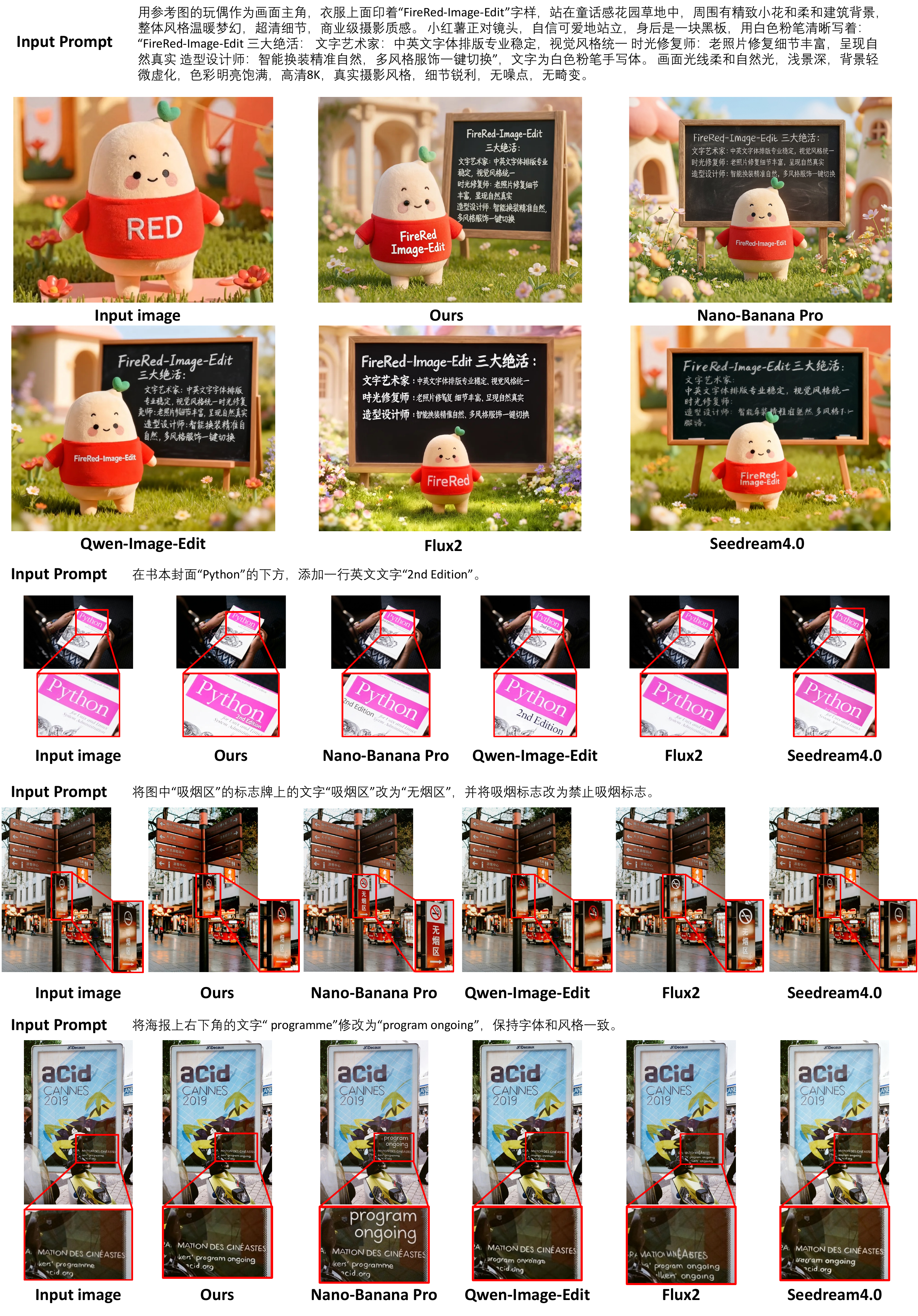}
    \caption{Text-centric showcases. Our model demonstrates superior legibility and style consistency for text editing. While competitors often struggle with blurred characters or layout distortion, our method achieves precise text rendering that preserves the original perspective and material texture.}
    \label{fig:text_showcase}
\end{figure}

\begin{figure}[H]
    \centering
    \includegraphics[width=\textwidth]{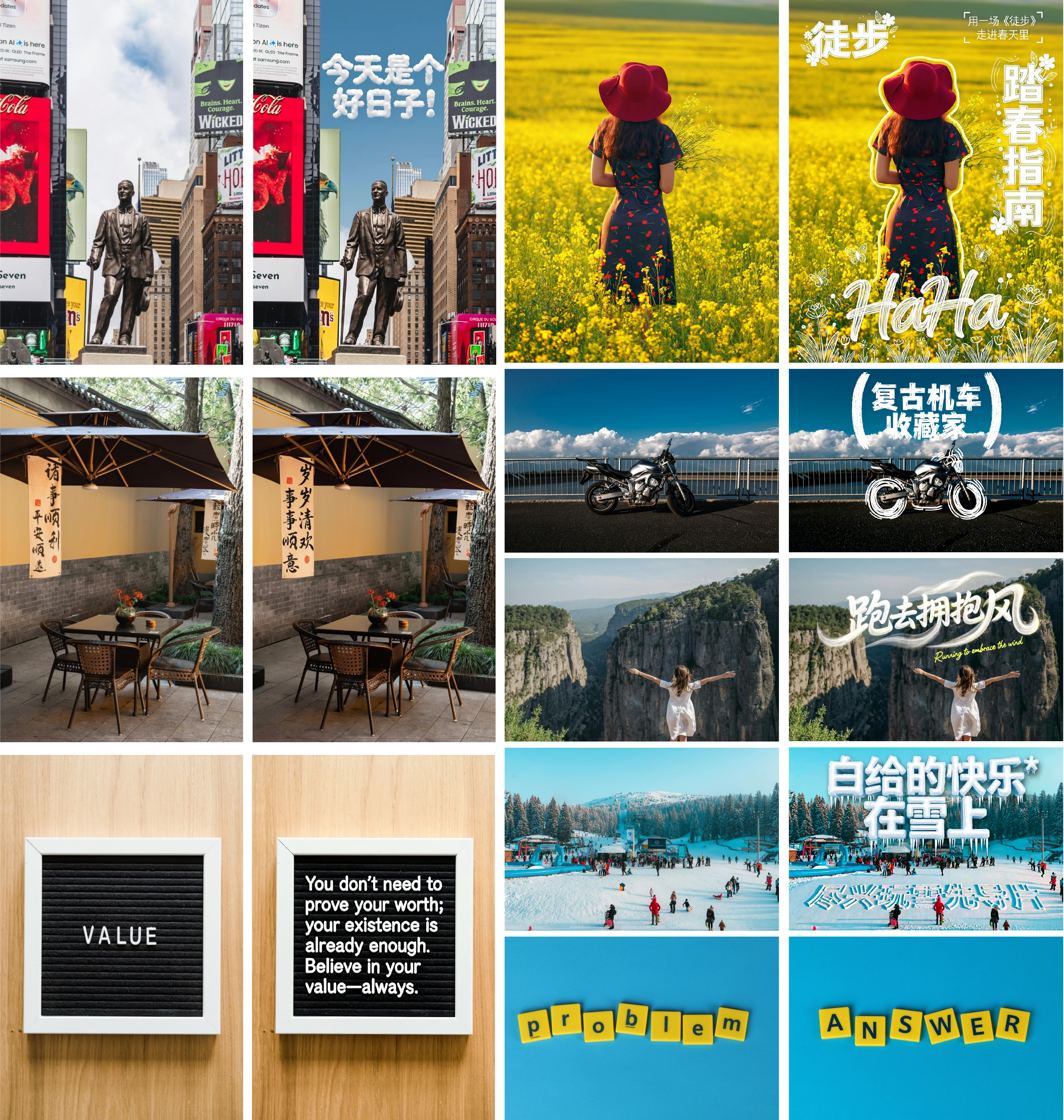}
    \caption{Text-centric showcases.}
    \label{fig:poster_showcase}
\end{figure}

\paragraph{Visualization.}
As illustrated in Fig.~\ref{fig:showcase_virtual_tryon}, each case involves combining the outfit from one image with a target model under additional constraints on garment fit (e.g., loose or tight silhouette), length, color specification, and accessory composition (e.g., handbags, jewelry, footwear). 
Such scenarios require accurate garment preservation, natural deformation under body pose, and consistent integration with newly introduced items, while maintaining identity and overall visual realism. 
Compared with existing editing baselines, \methodname produces more coherent clothing geometry, cleaner boundary transitions, and better alignment between textual styling instructions and visual outcomes. 
These results indicate that the model can handle compositional multi-constraint editing in a stable and controllable manner, making it suitable for practical fashion and styling applications.


\textit{}

\section{Conclusion}
\label{sec:conclusion}
We present FireRed-Image-Edit, a diffusion transformer for instruction-based image editing, with contributions spanning data engineering, training methodology, and evaluation.

In data engineering, we curate approximately one billion multi-source image–text pairs, covering text-to-image generation, multi-image synthesis, and diverse editing tasks. Through an end-to-end pipeline integrating rigorous cleaning, fine-grained stratification, auto-labeling, and a two-stage filtering mechanism, we distill over 100 million high-quality samples evenly balanced between generation and editing. This large-scale, structured curation substantially improves semantic alignment and data reliability.

In training methodology, we develop an efficiency-oriented framework featuring a Multi-Condition Aware Bucket Sampler to reduce padding overhead and Stochastic Instruction Alignment to enhance robustness. A multi-stage training pipeline (pre-training, supervised fine-tuning, reinforcement learning) incorporates Asymmetric Gradient Optimization for DPO to stabilize optimization. Additional components—including Multi-Image DPO, Diffusion NFT for text editing, Consistency Loss, Distributed Stratified Timestep Sampling, Logit-Normal loss, and EMA—jointly improve consistency, controllability, and convergence stability.

In evaluation, we introduce REDEdit-Bench, a benchmark spanning 15 structured editing categories with 1,673 bilingual samples, including newly designed portrait beautification and low-resolution enhancement tasks. Extensive experiments on REDEdit-Bench, ImgEdit, and GEdit demonstrate state-of-the-art performance among open-source models and competitive results against proprietary systems, validating that carefully engineered efficiency and system-level optimization can rival brute-force scaling.

\section{Authors}
\begin{itemize}
   \item \textbf{Core Contributors (listed alphabetically):}  Changhao Qiao, Chao Hui, Chen Li, Cunzheng Wang, Dejia Song, Jiale Zhang, Jing Li, Qiang Xiang, Runqi Wang, Shuang Sun, Wei Zhu, Xu Tang, Yao Hu, Yibo Chen, Yuhao Huang, Yuxuan Duan, Zhiyi Chen, Ziyuan Guo
    \item \textbf{Contributors(listed alphabetically):} Haohua Chen, Haolu Liu, Honghao Cai, Qing Yu, Shurui Shi, Shuyang Lin, Sijie Xu, Tianshuo Yuan, Tianze Zhou, Wenxin Yu, Xiangyuan Wang, Xudong Zhou, Xuecan Wang, Yahui Wang, Yandong Guan, Yanqin Chen, Yilian Zhong, Ying Li, Yunfan Liu, Yunhao Bai, Yushun Fang, Zeming Liu, Zhangyu Lai, Zhiqiang Wu
\end{itemize}

\clearpage

\addcontentsline{toc}{section}{References}
\bibliography{rededit_report}
\bibliographystyle{plain}

\appendix

\begin{promptbox}[Annotation Prompt]
\label{pb:Annotation Prompt}
\# Image Annotator\\
You are a professional image annotator. Please complete the following tasks based on the input image.

\vspace{1em}
\#\# Step 1: Image Caption
\begin{enumerate}
   \setlength{\itemsep}{0pt} \setlength{\parskip}{0pt} \setlength{\parsep}{0pt}
   \item Use natural, descriptive language without structured formats.
   \item Include details: object attributes, spatial relationships, and environment.
   \item Transcribe visible text exactly as shown, enclosed in quotation marks.
   \item Be specific and accurate, avoiding vague descriptions.
\end{enumerate}

\vspace{0.5em}
\#\# Step 2: Image Analysis
\begin{enumerate}
   \setlength{\itemsep}{0pt} \setlength{\parskip}{0pt} \setlength{\parsep}{0pt}
   \item Image Type: Classify by source (e.g., Photograph, Screenshot, Illustration).
   \item Visual Style: Identify artistic style (e.g., Photorealistic, Anime, Watercolor).
   \item Watermarks: List any detected watermarks, or "None".
   \item Anomalies: Note distracting elements like QR codes, mosaics, or artifacts.
\end{enumerate}

\vspace{0.5em}
\#\# Output Format\\
\verb|```json|\\
\verb|{|\\
\verb|  "caption": "Descriptive text with 'OCR text' quoted.",|\\
\verb|  "image_type": "Photograph",|\\
\verb|  "style": "Photorealistic",|\\
\verb|  "anomalies": ["QR code"] or [],|\\
\verb|  "subjects": {|\\
\verb|    "main": [|\\
\verb|      {|\\
\verb|        "appearance": "Age, gender, hair, build",|\\
\verb|        "clothing": "Type, color, accessories",|\\
\verb|        "action": "Pose or activity",|\\
\verb|        "emotion": "Expression and mood"|\\
\verb|      }|\\
\verb|    ],|\\
\verb|    "other": []|\\
\verb|  },|\\
\verb|  "environment": {|\\
\verb|    "setting": "Background elements, weather, season",|\\
\verb|    "lighting": "Light source and quality",|\\
\verb|    "palette": "Dominant colors",|\\
\verb|    "shot": "Camera angle and framing"|\\
\verb|  },|\\
\verb|  "quality": {|\\
\verb|    "resolution": "Sharp or blurry",|\\
\verb|    "watermarks": "Yes or No"|\\
\verb|  }|\\
\verb|}|\\
\verb|```|
\end{promptbox}

\end{document}